\documentclass[12pt]{elsarticle}

\usepackage[linesnumbered,algoruled,boxed,lined]{algorithm2e}
\usepackage{amsmath,amssymb,amsfonts}
\usepackage{array}
\usepackage{CJK}
\usepackage{color}
\usepackage{diagbox}
\usepackage{epsfig}
\usepackage{graphicx}
\usepackage{mathtools}
\usepackage{multirow} 
\usepackage[normalem]{ulem}
\usepackage{stmaryrd}
\usepackage{slashbox}
\usepackage{subcaption}
\usepackage{textcomp}
\usepackage{times}
\usepackage{url}
\usepackage{xcolor}
\usepackage{booktabs}
\usepackage{threeparttable}
\usepackage{tabularx}

\usepackage{hyperref}
\usepackage{color}
\usepackage[normalem]{ulem}
\newcounter{ToDo}
\newcounter{gaocomm}
\newcounter{Note}
\definecolor{blue-violet}{rgb}{0.54, 0.17, 0.89}
\definecolor{mygreen}{rgb}{0.0, 0.5, 0.0}
\definecolor{awesome}{rgb}{1.0, 0.13, 0.32}
\definecolor{bostonuniversityred}{rgb}{0.8, 0.0, 0.0}
\newcounter{mingcomm}
\definecolor{mycyan}{rgb}{1.00,0.00,0.50}
\definecolor{ao}{rgb}{0.0, 0.0, 1.0}
\definecolor{mygreen}{rgb}{0.0, 0.5, 0.0}
\definecolor{awesome}{rgb}{1.0, 0.13, 0.32}
\definecolor{bostonuniversityred}{rgb}{0.8, 0.0, 0.0}

\usepackage{hyperref}
\newcounter{wangcomm}
\definecolor{dark-red}{rgb}{0.7461,0.0234,0.1328}
\definecolor{mygreen}{rgb}{0.0, 0.5, 0.0}
\definecolor{awesome}{rgb}{1.0, 0.13, 0.32}
\definecolor{bostonuniversityred}{rgb}{0.8, 0.0, 0.0}

\newcounter{bxincomm}
\definecolor{aqua}{rgb}{0.00,0.67,0.80}

\usepackage{amsmath,amsfonts,bm}

\newcommand{\gph}{\mathcal{G}}
\newcommand{\tv}[2][j]{\mathcal{T}_{#2}^{(#1)}}

\def\eqref#1{eq.~(\ref{#1})}
\def\Eqref#1{Eq.~(\ref{#1})}
\def\1{\bm{1}}

\DeclareMathAlphabet{\mathsfit}{\encodingdefault}{\sfdefault}{m}{sl}
\SetMathAlphabet{\mathsfit}{bold}{\encodingdefault}{\sfdefault}{bx}{n}

\newcommand{\R}{\mathbb{R}}

\graphicspath{{./image/}}
\usepackage{placeins}

\usepackage{geometry}
\journal{Pattern Recognition}

\makeatletter
\def\ps@pprintTitle{%
 \let\@oddhead\@empty
 \let\@evenhead\@empty
 \def\@oddfoot{}%
 \let\@evenfoot\@oddfoot}
\makeatother

\begin{document}

\begin{frontmatter}

\title{\textsc{MathNet}: Haar-Like Wavelet Multiresolution-Analysis for\\ Graph Representation and Learning}
%{Graph Neural Networks with Multiresolution Wavelet Transforms}

%% Group authors per affiliation
\author[1]{Xuebin~Zheng\fnref{footnote1}}
\author[1]{Bingxin~Zhou\fnref{footnote1}\corref{correspondingAuthor}}\ead{bzho3923@uni.sydney.edu.au}
\author[2]{Ming~Li}
\author[3,4]{Yu~Guang~Wang}
\author[1]{Junbin~Gao}

\address[1]{The University of Sydney Business School, The University of Sydney, NSW 2006, Australia}
\address[2]{Department of Educational Technology, Zhejiang Normal University, Jinhua, Zhejiang 321004, P. R. China}
\address[3]{Max Planck Institute for Mathematics in the Sciences, Leipzig  04103, Germany}
\address[4]{School of Mathematics and Statistics, University of New South Wales, NSW 2052, Australia}
\fntext[footnote1]{Equal first contribution authors}
\cortext[correspondingAuthor]{Corresponding author}

\begin{abstract}
Graph Neural Networks (GNNs) have recently caught great attention and achieved significant progress in graph-level applications. In this paper, we propose a framework for graph neural networks with multiresolution Haar-like wavelets, or \textsc{MathNet}, with interrelated convolution and pooling strategies. The underlying method takes graphs in different structures as input and assembles consistent graph representations for readout layers, which then accomplishes label prediction. To achieve this, the multiresolution graph representations are first constructed and fed into graph convolutional layers for processing. The hierarchical graph pooling layers are then involved to downsample graph resolution while simultaneously remove redundancy within graph signals. The whole workflow could be formed with a multi-level graph analysis, which not only helps embed the intrinsic topological information of each graph into the GNN, but also supports fast computation of forward and adjoint graph transforms. We show by extensive experiments that the proposed framework obtains notable accuracy gains on graph classification and regression tasks with performance stability. The proposed \textsc{MathNet} outperforms various existing GNN models, especially on big data sets.
\end{abstract}


\end{frontmatter}

% \linenumbers

\section{Introduction}
%\GaoC{Can you split the references under each mentioned research area?}. 
Graph representation learning has broad applications in social network \cite{ying2018graph,perozzi2014deepwalk}, natural science \cite{hamilton2017representation,battaglia2018relational,ji2020action} and computer vision \cite{teney2017graph,wang2019dynamic,bear2020learning}. Recently, graph neural networks (GNNs) have been proved as an effective tool for structured data representation and learning. They have become a topic of intense research due to their remarkable ability on graph data modeling tasks, including node classification, graph classification and regression, and link prediction \cite{micheli2009neural,scarselli2008graph,scarselli2008computational,bronstein2017geometric,zhou2018graph,zhang2018deep,wu2020comprehensive}. %\GaoC{Can you split the references under each mentioned topic? Also checck others} 
In a GNN model, appropriate design of graph convolutional layers and pooling layers is crucial for the model to achieve outstanding performance. In this paper, we develop graph convolution and pooling operations based on Haar-like wavelets on graphs, which can achieve a state-of-the-art performance on graph classification and regression tasks.

\begin{figure}[t]
  \centering
  \includegraphics[width=.85\textwidth]{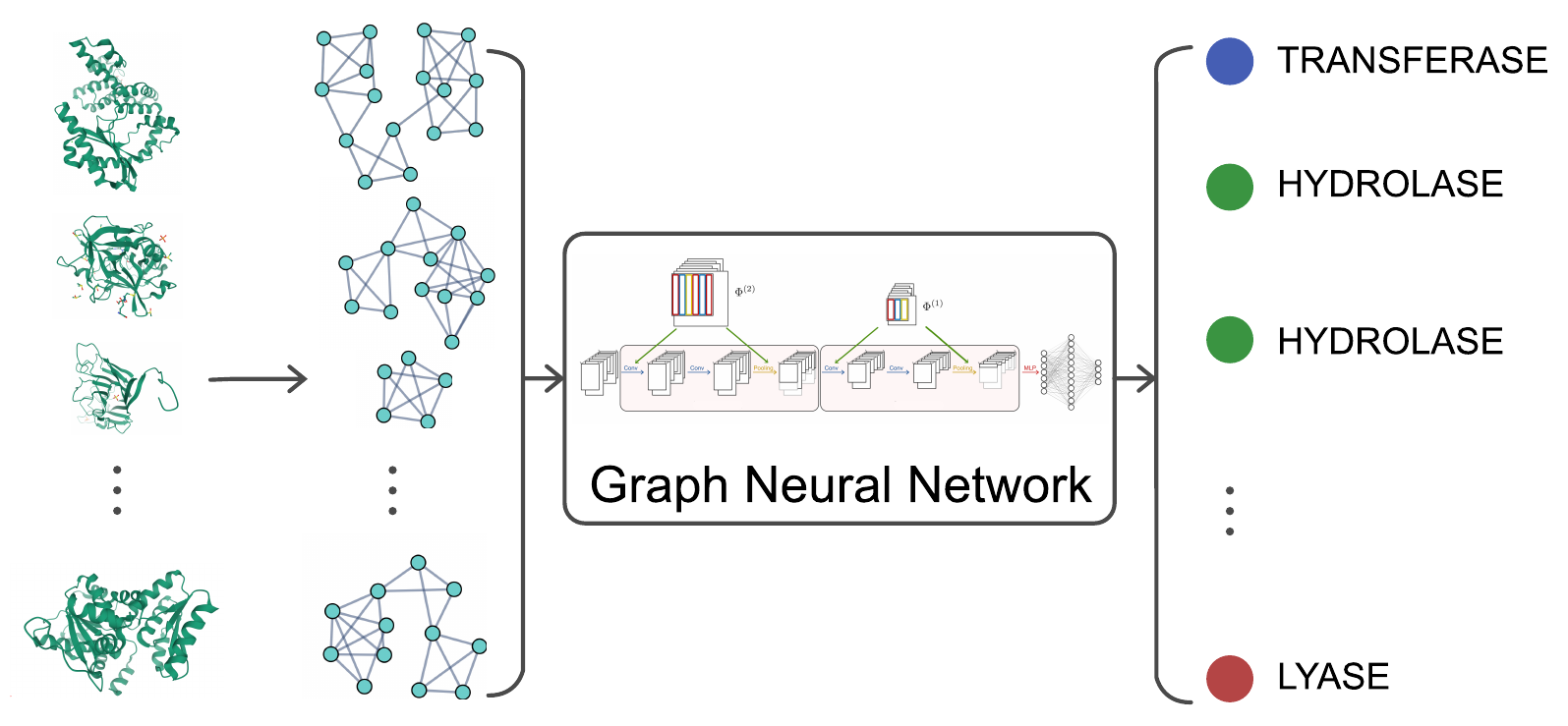}
  \caption{An example of applying GNNs for labeling enzymes. A set of graphs is extracted as the input data set, where each graph represents a specific protein tertiary structure. The task is to assign each enzyme instance to one of the given EC top-level classes correctly. The GNN's role is to find universally applicable rules to label the graphs by learning the topological and feature information of the input. \textsc{MathNet} provides a set of candidate GNN models with great potential. The structure of enzymes is retrieved from the Protein Data Bank \cite{pdb2000}.}\label{fig0}
\end{figure}

Node classification and graph classification/regression are typical tasks on graph-structured data. The node classification works on a single input graph which is to predict unseen labels on nodes from labeled nodes. Some applications include relation inference \cite{zhang2018link,kipf2018neural} and drug repurposing \cite{KnowledgeGraphCOVID19}. Graph classification or regression is the task of predicting unknown labels of individual input graphs by learning from multiple labeled graphs. The size and structure of input graphs are possibly distinct from one to another. The applications of graph classification/regression include protein structure classification \cite{jiang2019walk}, point distribution pattern recognition in statistical physics \cite{ma2019pan,ma2020path} and atomization energy prediction in quantum chemistry \cite{gilmer2017neural,wu2018moleculenet}.

Graph neural networks are useful for solving these problems. GNNs are deep neural networks with general architecture analogous to traditional CNNs except the input is graph-structured data. This difference, however, has a big impact on the construction of the network and its components, such as the counterparts of traditional convolution and pooling. In a GNN, graph convolution is an operation that can distill a geometric feature of the structured data which applies to both node-level and graph-level tasks. One typical graph convolution with discrete Fourier transform carries computations in Fourier domain: the input graph features and network filter are projected from the vertex domain by the forward wavelet transform, and the convolution is then computed by the Hadamard product of the projections \cite{bruna2013spectral}. The adjoint Fourier transform finally takes the processed signal back to the vertex domain as the output. This spectral-based graph convolution alone, like other graph convolutions, cannot deal with graph-level classification when the input graph has a different size and structure in the user-specified network architecture. A common solution is to insert a graph pooling layer between graph convolution and the final readout layer \cite{ying2018hierarchical,lee2019self}. The target of graph pooling operation is to guarantee that the output as a low-dimensional representation of the graph data has a unified size and can be further analyzed by readout layers such as a fully-connected layer.

To this end, we develop a wavelet version of graph convolution and graph pooling with a class of discrete orthonormal Haar-like wavelets. We name the proposed framework \textsc{MathNet}, which is short for Graph Neural networks of \textbf{M}ultiresolution-\textbf{A}nalysis \textbf{T}ransforms based on \textbf{H}aar-like Wavelets. The selected wavelets are fully compact with the multiresolution analysis (MRA), which decomposes data into different scales then conduct analysis. Given a set of input graphs, the wavelet-based MRA for each graph could be constructed from a coarse-grained chain of the graph. The orthonormal system is built upon this chain, which can later guide fast forward and adjoint wavelet transforms for both graph convolution and pooling operations. Meanwhile, this tree-structured system also supports a cascading computing strategy for the pooling layer, which makes it a hierarchical design with multiple pooling layers. Each pooling layer compresses the node number by combining the feature values and graph topology of the current layer's input. The implementation of this wavelet-based graph pooling results in a compressive wavelet transform of the input features.

The wavelet-based MRA methods bring several merits to graph representation learning tasks. It now becomes possible to decompose and process information at different scales. Both local and global information of graph data are extracted from the hierarchical orthonormal system, and both the topological and feature values are taken care of by the wavelet transform. Besides, the established sparse orthonormal system supports fast computation of forward and adjoint transforms in graph convolution layers. In pooling layers, the wavelet compression cuts down node numbers without sacrificing the major geometric or feature information of the input graph. 

To the best of our knowledge, this is the first work to incorporate Haar-like MRA wavelets with convolution and pooling layers in one GNN for graph properties prediction tasks. The proposed framework is not only efficient of solving large-scale node-level learning problems, but also outstanding in handling graph-level problems. Aside from the theoretical merit of the proposed framework, we provide extensive empirical evidence to support that our proposed \textsc{MathNet} can outperform baseline methods in many scenarios.
%instruction of constructing a general multiresolution wavelet based graph neural network, we also provide a specific example with Haar-like wavelets, of which the empirical evidence supports the superiority of our proposed framework.

The rest of this paper is organized as follows. Section~\ref{relatedWork} reviews classic works of learning graph representation as well as recent major developments. Section~\ref{preliminary} introduces the fundamental ideas and notions of discrete wavelets and MRA. We also present a discussion on the rationality of building graph representation with Haar-like wavelets. The proposed graph neural networks framework \textsc{MathNet} is then introduced in Section~\ref{haarSystem}, where we give full details from constructing the global orthonormal basis to fast computation that supports graph convolution and graph pooling operations. In Section~\ref{experiment}, we test $\textsc{MathNet}$ on a variety of graph learning tasks, including both classification and regression tasks on benchmark and new data sets. Section~\ref{conclusion} summarizes the paper together with a discussion on future work.

\section{Related Works}
\label{relatedWork}
Researchers have probed approaches for building graph convolutional layers. Two representative types are the spectral-based method and spatial-based method. The spectral graph convolution, introduced in \cite{bruna2013spectral}, is based on convolution theorem and spectral graph theory \cite{chung1997spectral}. The spectral graph convolution is realised based on the graph Fourier transforms, which suffers from the high computational complexity of the Laplacian eigendecomposition; and there is, in general, no fast implementation for graph Fourier transforms due to the non-Euclidean structure of graph data. 

On the other hand, filters in the Fourier domain cannot guarantee the localisation in the spatial (vertex) domain. ChebNet proposed in \cite{defferrard2016convolutional} uses Chebyshev polynomial approximation for graph convolution, which constructs a localised polynomial filter, and circumvents the computation of the Laplacian eigendecomposition. GCN \cite{kipf2016semi} simplifies ChebNet by using the first order Chebyshev polynomial of the graph Laplacian with renormalisation tricks in computation. Authors of \cite{xu2018graph,li2019haar,ma2019multi,wang2019haarpooling,zheng2020decimated} developed wavelet-based GNNs by replacing graph Fourier transforms with graph wavelet transforms. Due to the high sparsity of wavelet basis matrix, the graph wavelet transforms are more computationally efficient than graph Fourier transforms. 

Compared with spectral-based graph convolutions, the spatial-based methods are closer to the conventional CNN convolution on image classification as the graph convolutions are computed based on a node's spatial relation and the learnable filters are defined in the vertex domain \cite{atwood2016diffusion,veli2018graph,xu2018how,hamilton2017inductive,monti2017geometric}. The spatial node information aggregation was first developed in \cite{micheli2009neural,scarselli2008graph} for recursive and recurrent neural networks.
In the domain of GNN, Gilmer et al. \cite{gilmer2017neural} provided the framework of message passing neural networks (MPNNs), encapsulating many existing GNNs in the viewpoint that information/message can be passed from one node to another along edges/paths directly.

While graph convolution layers aim to extract the high-level node representation, graph pooling layers are needed to obtain the graph-level representations necessary for graph classification and regression problems as concerned in this work. The general process is to coarsen a graph into subgraphs, so that node representations on coarsened graphs have higher graph-level representation. A readout layer can then be used to incorporate the node representation of each graph into a unified graph representation. Authors of \cite{gilmer2017neural,atwood2016diffusion,duvenaud2015convolutional} extended the global sum/average pooling operation to graph models by summing or averaging all node features. Some advanced graph pooling methods, such as DiffPool \cite{ying2018hierarchical}, SortPool \cite{zhang2018end}, TopKPool \cite{gao2019graph,cangea2018towards}, SAGPool \cite{lee2019self}, are proposed with special consideration on the hierarchical representation and feature information.  Apart from that, Noutahi \emph{et al} \cite{noutahi2019towards} and Ma \emph{et al} \cite{ma2019graph} took the graph Laplacian into account to combine the feature information and the structural information. Such spectral-based techniques drop the detailed information layer by layer to extract the smoothing feature representation. Unfortunately, both methods have high computational cost mainly due to the Laplacian eigendecomposition.
\section{From Wavelet Multiresolution Analysis to Graph Learning}
\label{preliminary}
In this preliminary section, we describe the notions and definitions for developing a Haar-like graph representation. We begin with formulating our objective to predict graph properties by analyzing the input graph features and structures. We then revisit the definition of discrete wavelet transforms that help process graph signals on a transformed vertex domain. The pyramid algorithm for discrete wavelets is then discussed in Section~\ref{mra}. This tree-shaped organization supports a sparse global orthonormal basis construction of input signal and backbones fast algorithms on the forward and backward wavelet transforms. Finally, to close this section, we list the rationale of introducing such multi-scale Haar-like wavelet system for graph representations.

\subsection{Notations}
The graph-level task is to predict for a specific graph $(\mathcal{G}_g, X_g)$ its label $\mathcal{Y}_g$, where $\mathcal{G}_g = (\mathcal{V}_g, \mathcal{E}_g, \mathcal{W}_g)$ is an undirected graph instance, and $X_g$ contains the node-level feature information. For the graph structure, we denote the $\mathcal{V}_g = \{v_{1}, v_{2},\dots, v_{N_g}\}$ a non-empty finite set of $|\mathcal{V}_g| = N_g$ vertices, and $\mathcal{E}_g \subset \mathcal{V}_g \times \mathcal{V}_g$ a set of edges that can be represented as a weighted symmetric adjacency matrix $\mathcal{W}_g \in \mathbb{R}^{N_g \times N_g}$. Then for a set of input graphs $\{(\mathcal{G}_1, X_1), (\mathcal{G}_2, X_2), \dots, (\mathcal{G}_M, X_M)\}$, the desired output is a set of labels $\{\mathcal{Y}_1, \mathcal{Y}_2, ..., \mathcal{Y}_M\}$.

We now omit the subscript $g$ and focus on a specific graph $\mathcal{G}$. A \emph{coarse-grained chain} describes a sequence of graphs
\begin{equation*}
\mathcal{G}^{J \rightarrow J_0} \coloneqq 
(\mathcal{G}^{(J)}, \mathcal{G}^{(J-1)}, ..., \mathcal{G}^{(J_0)}) 
\end{equation*} 
where $J > J_0$.
This tree representation is usually achieved by progressively coarsening the graph into smaller partitions. The $\mathcal{G}^{(J)}$ is \textit{the finest level} or \textit{the bottom level}, which denotes the original graph. Any member at the layer that is coarser than the finest layer is a \emph{coarse-grained graph} of its finer layer(s), and the last graph $\mathcal{G}^{(J_0)}$ is recognized as \textit{the coarsest level} or \textit{the top level}. Further, the graph nodes from a coarser layer are called \emph{parents}, and the nodes from any finer layer(s) are called \emph{children}. At level $j$, the set of vertices is denoted by
\begin{equation*}
\mathcal{V}^{(j)} = \{v_{1}^{(j)}, ..., v_{N^{(j)}}^{(j)}\},
\end{equation*}
where $N^{(j)} \coloneqq |\mathcal{V}^{(j)}|$ indicates the number of vertices in graph $\mathcal{G}^{(j)}$. On top of the aforementioned graph symbols, we also list pivotal notations in Table~\ref{notation} so that our work could be easier to follow.

\begin{table}[th]
\caption{Table of Notations.}\vspace{-5mm}
\begin{center}
\begin{tabular}{l l}
\toprule
\textbf{Notation} & \textbf{Description}\\
\midrule
$\mathcal{G}^{(j)}$ & A subspace of the raw graph $\mathcal{G}$ at the $j$-th level\\
%$\widetilde{\mathcal{G}^{(j)}}$ & The orthonormal complement of $\mathcal{G}^{(j)}$ to $\mathcal{G}^{(j+1)}$\\
$\mathcal{G}^{J \rightarrow J_0}$   & A coarse-grained chain of graphs $(\mathcal{G}^{(J)}, \mathcal{G}^{(J-1)}, \dots, \mathcal{G}^{(J_0)}) $ \\
$\mathcal{V}^{(j)}$ & A non-empty finite set of vertices $\{v_{1}^{(j)}, v_{2}^{(j)}, ..., v_{N^{(j)}}^{(j)}\}$ in $\mathcal{G}^{(j)}$\\
$\mathcal{W}^{(j)}$ & The weighted adjacency matrix for the $j$th coarse-grained graph\\
$N^{(j)}$   & The number of vertices in $\mathcal{V}^{(j)}$\\
$n_{\ell}^{(j)}$    & The number of vertices in the $\ell$th cluster of in $\mathcal{G}^{(j)}$\\
$v$ & An arbitrary vertex at a finer level of the graph\\
$[v]$ & The cluster that contains $v$, or the parent $[v]$ as a vertex of a coarser level\\
$v_{\ell,k}^{(j)}$ & The $\ell$th vertex from the $k$th cluster of level $j$\\
%$\{\Phi^{(J_0)},\dots,\Phi^{(J)}\}$ & An orthonormal basis system of $\mathcal{G}^{J \rightarrow J_0}$\\
$\phi_{\ell}^{(j)}(v)$  & The $\ell$th basis of $\Phi^{(j)}$ at level $j$\\
$\phi_{\ell}^{(j)*}(v)$ & A vertically extended basis of length $N^{(j)}$ from $\phi_{\ell}^{(j)}(v)$ \\
$c_{\ell}^{(j)}$    & The $\ell$th wavelet coefficient at level $j$ or the wavelet coefficient of $v_{\ell}^{(j)}$\\
$S_{v}^{(j)}(x)$   & The summation for $x$ of the cluster $[v]\in\mathcal{V}^{(j-1)}$ in the forward wavelet transform\\
$S_{v}^{(j)}(c)$   & The summation for $c$ of the vertex $v\in\mathcal{V}^{(j)}$ in the adjoint wavelet transform\\
$G$ & Trainable filter matrix in the wavelet domain\\
$\widetilde{\Phi}^{(j)}$ & The pooling operator with the first $N^{(j-1)}$ basis vectors of $\Phi^{(j)}$\\
\bottomrule
\end{tabular}
\label{notation}
\end{center}
\end{table}

\subsection{Discrete Wavelet Transform}
We call a function $\phi(v)$ a \emph{wavelet} if it yields an orthonormal basis 
\begin{equation}
\left\{\phi_{a,b}(v) = C \cdot \phi\left(\frac{v-b}{a}\right): a>0 \text{ and } b\in\mathbb{R}\right\}
%\phi_{j,n}(x)=2^{-j/2}\phi(2^{-j}x-n)
%\phi_{a,b}(x)=|a|^{-1/2}\phi(\frac{x-b}{j}),  a,b\in\mathbb{R}, j\neq 0
\label{waveletBasis}
\end{equation}
in $L^2(\mathbb{R})$ by translation and dilation of itself. Under this construction, any elements in $L^2(\mathbb{R})$ can be represented as a linear combination of the basis. The two parameters $a$ and $b$ indicate the underlying signal $\phi(v)$ being dilated at the scale $a$ and translated to the position $b$, and $C$ is a constant scalar with respect to the scale $a$. Specifically, when $a=2^j$, it comes to the classical dyadic dilation. The basis function $\phi(v)$, also called the mother wavelet, is chosen to serve as a prototype for all basis functions in the process. For example, in the Haar-like wavelet basis (as we shall introduce below) designs $\phi(v)$ as a characteristic function of the interval $[0,1)$. Each function in the orthonormal basis, $\phi_{a,b}(v)$, defines a wavelet transform that encodes the signal $x = f(v) \in L^2(\mathbb{R})$ 
from time domain to frequency domain. This transform aims at representing the signal $x$ by different frequency components, each with a specific scale. (So one can extract information from the original signal at a coarse or fine scale.) In comparison to the conventional Fourier transforms, the wavelet transforms expand the scaling function beyond the exponential family and obtain both time and frequency localizations, which makes it capable for non-stationary signal processing. 

An eligible wavelet function is generally restricted to be admissible and regularized \cite{medhat2004review}. The first requirement is usually called the \emph{admissibility condition}. It explains the function’s name of `wave’, which requires the function to be oscillating and finite to guarantee a stable reconstruction of signal \cite{valens1999really}. The second condition of \emph{regularization} is closely related to the number of vanishing moments that control the wavelet function's smoothness. An increasing number of vanishing moments provide smoother wavelets or scaling functions, making the wavelet bases decrease quickly with decreasing scales and helps the transform distinguish the essential information of signal from non-essential information and noise. 

Like the Fourier counterparts, the wavelet transform can be distinguished between the continuous and the discrete wavelet transform. Depending on the construction details, the discrete wavelet transform system could further be torn down to redundant frames, orthonormal bases, or other categories \cite{mallat1999wavelet}. Since graph data naturally have a discrete structure, we will develop discrete wavelet transforms in this work. While we consider systems without redundancy as a starting point, generalizing the research to redundant frame systems and enabling extra flexibility is possible in the future. To this end, we aim at constructing a wavelet-like representation of graph functions with filtration, that is, a mechanism to describe the `parent-children' relationships of the instances \cite{chui2015representation}. 

Our method dilates and translates signal information along with a cascade tree structure of graph data. In literature, the construction result is usually called a multiresolution orthonormal wavelet system. If we further restrict the mother wavelet to be piece-wise constant, it becomes tree-based Haar-like wavelets. There are many desired properties of Haar-like wavelets. For example, it has a small vanishing moment of 1 with the shortest support among all orthogonal wavelets. It promises a stationary transformation of localized data. As a result, data from different regions would not affect each other during the wavelet transform. It is also the only orthonormal and symmetric wavelet that is compactly supported \citep{daubechies1988orthonormal}. Although the discrete properties weaken Haar-like wavelet's ability to approximate smooth functions, its simplicity and formidable applicability assign it tremendous educational value and make it possible to be generalized to many sophisticated scenarios. 

In the next section, we will introduce the tree-based wavelet basis construction. Rather than producing graph representation with eigenvalues and eigenvectors of the graph Laplacian, this multi-scale method generates a pyramid of sub-trees that describe local relationships of graph vertices. The consequent orthonormal system from this pyramid-based construction is sparse and fully supported by fast computation algorithms, see Section~\ref{haarSystem}.

\subsection{Multiresolution Analysis}
\label{mra}
The idea of multiresolution analysis (MRA) was first linked to wavelets and cascade algorithms in \cite{mallat1989theory}. The motivation of MRA is to approximate $L^2(\mathcal{G}^{(J)})$ as the union of a sequence of bases on the subspaces $\mathcal{G}^{(j)}$ of $\mathcal{G}^{(J)}$, $j \in \mathbb{Z}_{\geq0}$, where all the subspaces are evolved from the central reference space $\mathcal{G}^{(J_0)}$, and each $\mathcal{G}^{(j)}$ is a subspace of $\mathcal{G}^{(j+1)}$. 
% We call the sequence $\{\mathcal{G}^{(J_0)},\dots,\mathcal{G}^{(J)}\}$ an MRA of $L^2(\mathcal{G}^{(J)})$.
A direct application of MRA is to construct orthonormal wavelet bases for the fast discrete wavelet transform algorithm. The expanded pyramid structure of the subspaces stores the details for raw data on an orthonormal basis, where lower-level subspaces preserve detailed information while upper-level subspaces contain approximation information. %For the rest of this section, we shall drop the subscript $b$ in $\phi$ and $\psi$ for the sake of simplicity. In practice, this $b$ will be determined by the number of vertices in the corresponding subgraph or subspace.

We first introduce the notion of a wavelet basis. We call $\{\Phi^{(J_0)},\dots,\Phi^{(J)}\}$ a \emph{scaling-like system} if each $\Phi^{(j)} = \{\phi_1^{(j)},\dots,\phi_{N_j}^{(j)}\}$ spans the subspace $L^2(\mathcal{G}^{(j)})$. We can further drop the redundancies and leave only orthogonal elements $\psi^{(j)}_{\ell}$ for some scale $j$ and position $\ell$, such that $\langle\psi^{(j)}_{\ell},\psi^{(j)}_{\ell'}\rangle=0$ when $\ell \neq \ell'$. The resulting $\{\Psi^{(J_0)},\dots,\Psi^{(J)}\}$ is a \emph{wavelet-like system} with respect to the scaling-like system, and the $\{\Phi^{(J_0)},\Psi^{(J_0)},\dots,\Psi^{(J)}\}$ is an \emph{orthonormal basis} for $L^2(\mathcal{G}^{(J)})$ or a wavelet basis.

Consider the approximation of $x$ at the scale $j$ as the orthogonal projection $\operatorname{Proj}_{\Phi^{(j)}}x$ on the space spanned by the basis $\Phi^{(j)}$ on the graph $\mathcal{G}^{(j)}$ of level $j$. The \emph{pyramid algorithm} \cite{mallat1989theory} suggests approximating on a fine scale by adding detail information to the approximation information on a coarse scale. For example, the approximation of $x$ on $\mathcal{G}^{(j+1)}$ is 
\begin{equation}
\operatorname{Proj}_{\Phi^{(j+1)}} x = \operatorname{Proj}_{\Phi^{(j)}} x + \operatorname{Proj}_{\Psi^{(j)}} x, 
\label{projection}
\end{equation}
where $\Psi^{(j)}$ is the orthonormal complement of $\Phi^{(j)}$ to $\Phi^{(j+1)}$, i.e., $\Phi^{(j+1)} = \Phi^{(j)} \oplus \Psi^{(j)}$ where $\oplus$ represents orthogonal sum. In this way, the raw signal $x$ is progressively smoothed by an iterative procedure, where each level $j+1$ constitutes the approximated information from the orthogonal projection $\operatorname{Proj}_{\Phi^{(j)}} x$ and the detailed information $\operatorname{Proj}_{\Psi^{(j)}} x$ that appears at the level $j+1$ but disappears at a coarser level $j$. 
% In other words, we split the space $L^2(\mathcal{G}^{(J)}) = \mathcal{G}^{(J_0)} \oplus \left(\oplus_{j=J_0+1}^{J} \widetilde{\mathcal{G}^{(j)}}\right)$. 
The above \Eqref{projection} from left to right 
provides decomposition and reconstruction for the wavelet representation of a graph signal.

For the reconstruction of the original signal $x$, we start by finding the orthogonal projection of $x$ on $\Psi^{(j)}$ with an expansion of the wavelet basis
$$
\operatorname{Proj}_{\Psi^{(j)}} x = \sum_{\ell=1}^{N^{(j)}}\langle x, \psi^{(j)}_{\ell}\rangle \psi^{(j)}_{\ell}.
$$
Follow the same procedure we aggregate all details at all scales plus the approximation at the coarsest level, and reconstruct the signal
\begin{align}
x = \operatorname{Proj}_{\Phi^{(J_0)}} x + \sum_{j=J_0}^{\infty}\operatorname{Proj}_{\Psi^{(j)}} x
%= \sum_{\ell=1}^{N^{(J_0)}}\langle x, \phi^{(J_0)}_{\ell}\rangle \phi^{(J_0)}_{\ell}
%+ \sum_{j=J_0}^{J}\sum_{\ell=1}^{N^{(j)}}\langle x, \psi^{(j)}_{ \ell}\rangle \psi^{(j)}_{\ell}
\label{reconstruction}
\end{align}
with both approximation information on $\Phi^{(J_0)}$ and detail information on $\Psi^{(j)}$, $J_0 \leq j \leq J$. The perfect reconstruction relies on the admissibility of the mother wavelet, which is a core requirement of wavelets. With this property, the wavelet coefficients could completely characterize $x$.

We have introduced the idea of wavelets and the key steps for decomposition and reconstruction. Next, we explain the merits of applying this system to graph representation.

\subsection{Bridge to Graph Representation}
The well-structured wavelets make it a useful tool for many applications such as signal processing and data mining. In this section, we present the main features of Haar-like MRA Wavelets that are desirable for graph representation learning tasks. 
\begin{enumerate}
    \item \textbf{Self-similarity and localization of neighbors}: The Haar-like wavelet bases is supported on a finite interval. This support property is essentially closely related to vanishing moments, which promises the oscillatory nature of wavelets and characterizes the differences between an object and its neighbors. That is, the wavelets are guaranteed to be localized, and objects from different regions would not affect each other. This property obeys the intuition of graph representation learning, where unconnected `strangers' are assumed to have little connections with each other.
    \item \textbf{Hierarchical Representations and Manipulations}: MRA provides an explicit method for constructing orthonormal wavelet bases from a scaling function. The signal on the wavelet domain is divided into `low-pass' and `high-pass' channels, where the high-pass noisy information usually corresponds to small wavelet coefficients and the wavelet coefficients of less noisy low-pass data can have a large variation. Aside from treating different levels of approximated and detailed signals separately in the graph convolution process by learning and endowing different weights, this difference is also a natural echoing to the popular hierarchical graph pooling methods. With wavelet compression, the preserved information tends to cover the majority information from the disposed small-valued noisy data.
    \item \textbf{Linear Computational Complexity}: Mallat's pyramid algorithm provides an efficient way for processing the Discrete Wavelet Transform. On the graph, the fast algorithm for Discrete Wavelet Transform can achieve $\mathcal{O}(n)$ computational complexity.
\end{enumerate}

\begin{figure*}
  \centering
  \includegraphics[width=\textwidth]{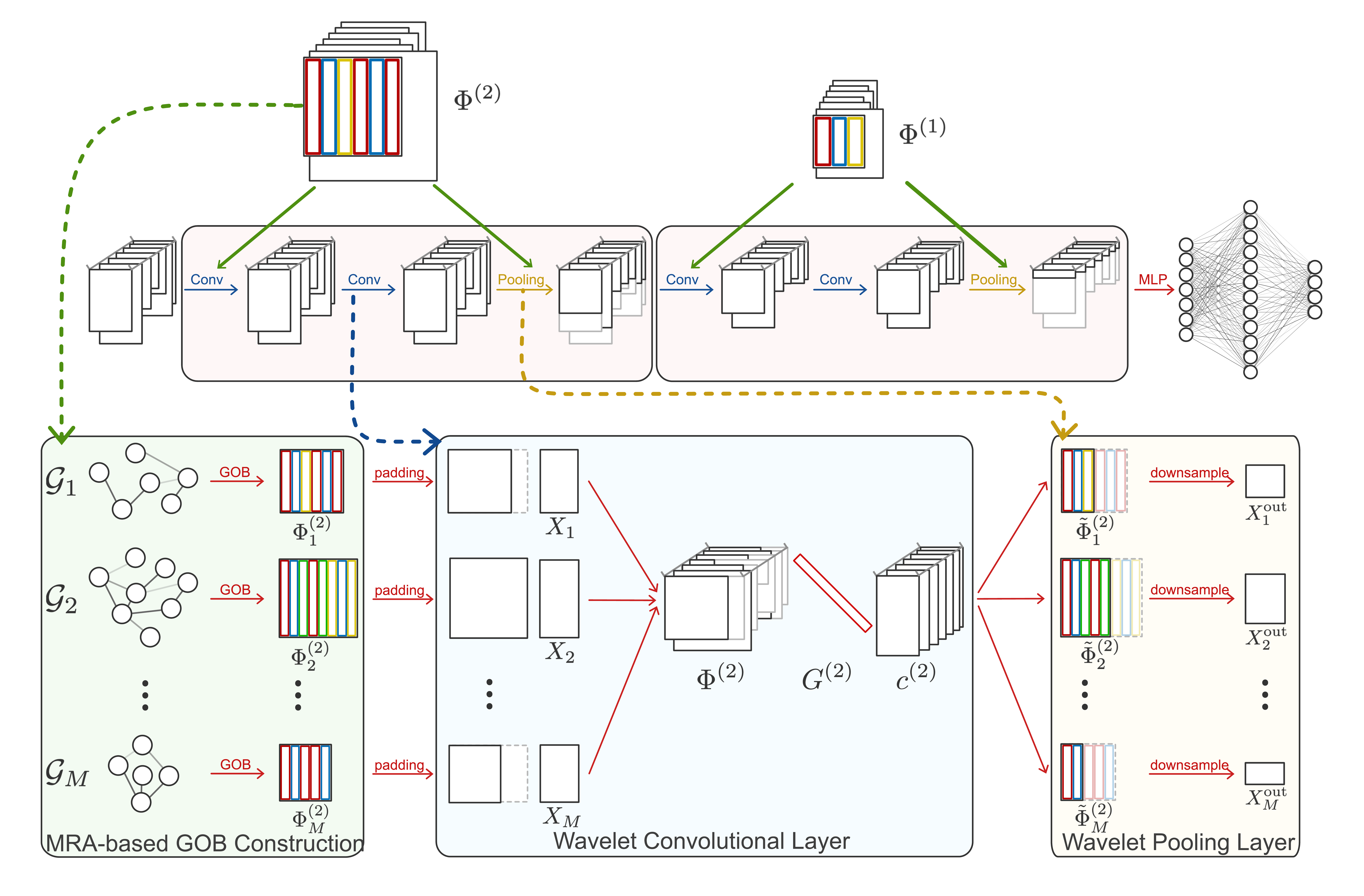}
  \caption{An illustration of the general $\textsc{MathNet}$ structure with 2 operational blocks and a fully connected layer. We unify a block of the $\textsc{MathNet}$ structure (red rounded-rectangles) to cover graph convolutional layer(s) following graph pooling layer(s). Given a set of graph $\{(\mathcal{G}_1, X_1), (\mathcal{G}_2, X_2), \dots, (\mathcal{G}_M, X_M)\}$, we first construct the wavelet basis sets $\Phi^{(2)}$ and $\Phi^{(1)}$ along the coarse-grained chain for each graph $\mathcal{G}_g^{(j)}$ at level $j = 2$ and $1$. The characterized orthonormal systems are used to support fast convolutional and pooling operations. The formulations inside a block only involve one basis set. In the first block, we take the ordered wavelet global orthonormal bases $\Phi^{(2)} = \{\Phi_1^{(2)}, \Phi_2^{(2)}, \dots, \Phi_M^{(2)}\}$ to guide the forward and adjoint discrete wavelet transforms. The process, as detailed in the blue rounded-rectangle, constitutes one wavelet graph convolution layer. The filter matrix $G$ is shared among graphs at the same level, and it is universally applicable on different graphs after the padding strategy on $\Phi$. For the sake of conciseness, we only demonstrate one convolution layer. This process, in practice, could repeat multiple times. We send the final output of the convolution layer (after activation) to the pooling layer, where we utilize the compressive wavelet transforms to filter out the detail information and reserve approximation information. The coarsening process is implemented by the transform of $\Phi^{(2)}$; the $\Phi^{(1)}$ (yellow rounded-rectangles) gives the size of the output of the downsampling procedure. Note that the network component is not constrained to the illustration, and a specific application should determine the employed structure. For clarity, this visualization excludes the common technical details like activation and weight detaching.}
  \label{fig3}
\end{figure*}
\section{Graph Convolution and Pooling by Haar-like Wavelet Transforms}
\label{haarSystem}
So far we assume a known local filtration with a given dilation ratio at each scale from a properly defined multiresolution view of the input. However, the nest structure could be less obvious when it comes to structured data like graphs. To this end, we start this section by introducing a coarse-grained chain algorithm that constructs a tree representation of graph data. We then present the general strategy for a graph properties prediction task, which requires two steps, including graph convolution and graph pooling. The convolution step first transforms the graph information to the wavelet domain and extracts useful patterns by a trainable filter $G$ (parameters). The processed signals are then decoded back to the original vertex domain. Besides, hierarchical graph pooling layers could be applied to coarsening the graph structure from different scales, and the compressed numeric representation of each graph instance is ready to be sent to the readout layer. A comprehensive description of the process is visualized in Figure~\ref{fig3}.
% In this section we will give the construction details of a specific multiresolution wavelet graph neural network $\textsc{MWGNN}_{\textsc{Haar}}$, of which the wavelet type is specified as Haar. As the simplest member of wavelet, the piece-wise constant function of Haar wavelet fit the nature of tree structure very well. We start from constructing the Haar Orthonormal System in Section~\ref{methodology_haarBasis} with basis extension and Gram-Schmidt orthonormalization. We then present the fast forward and adjoint Haar transform for graph convolution layers of $\textsc{MWGNN}_{\textsc{Haar}}$ following with the padding strategy to have consistent filters among graphs. In graph pooling layers, we conduct a similar procedure as the fast adjoint transform and produce the compressive Haar transform. 

\subsection{Coarse-grained Chain Encoder}
\label{CGC}
\begin{figure}
  \centering
  \begin{minipage}[t]{0.48\textwidth}
    \includegraphics[width=\textwidth]{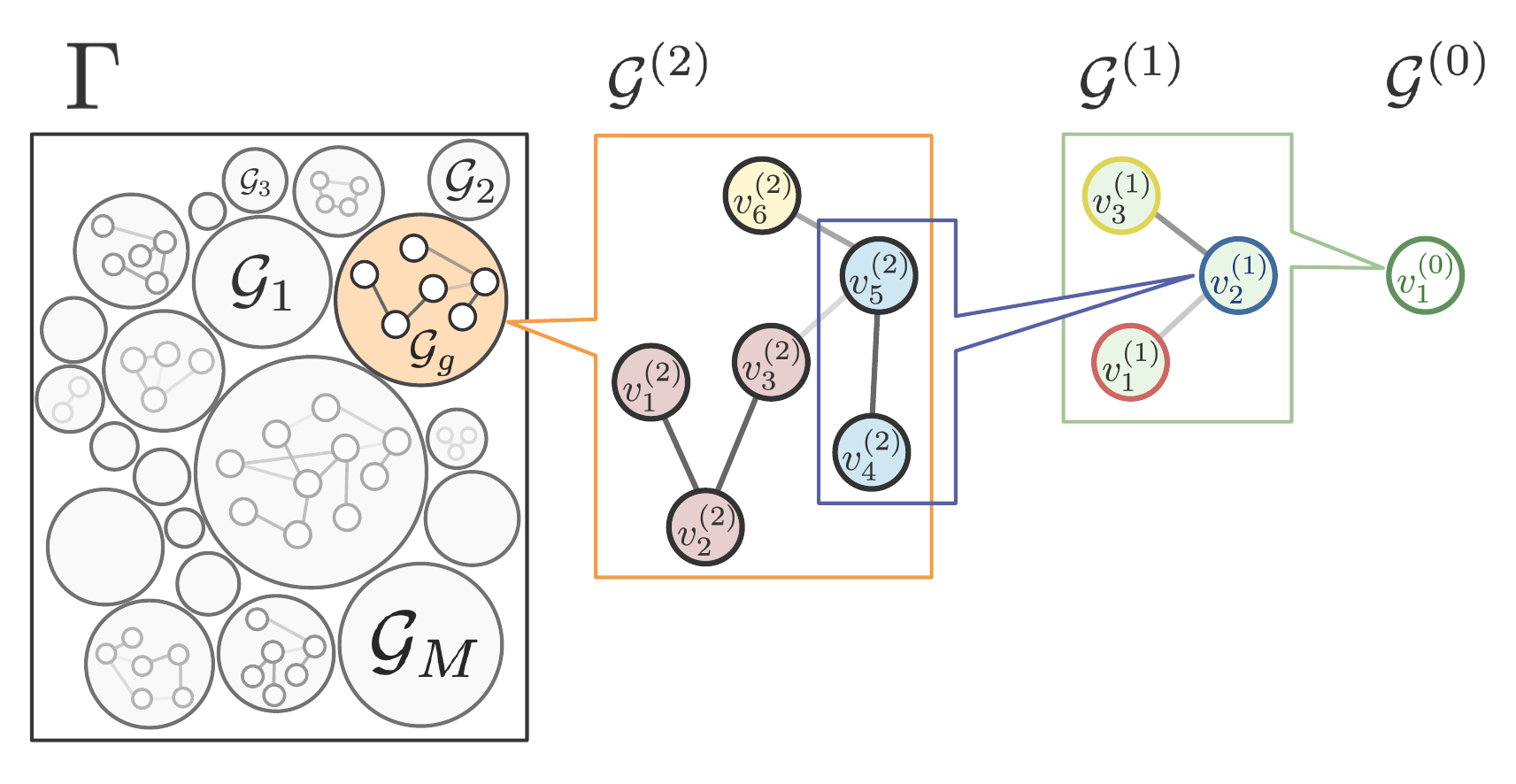}
    \caption{The hierarchical relationship among a graph instance $\mathcal{G}_g=\mathcal{G}^{(2)}$ and its coarse-grained graph $\mathcal{G}^{(j)}$ with $j = 0, 1$. The level-wise orthonormal bases for every graph instance are generated with the constructed chain, where the orthonormal bases guide the corresponding graph convolution and graph pooling operations.}
    \label{fig1}
  \end{minipage}
  \hfill
  \begin{minipage}[t]{0.48\textwidth}
    \includegraphics[width=\textwidth]{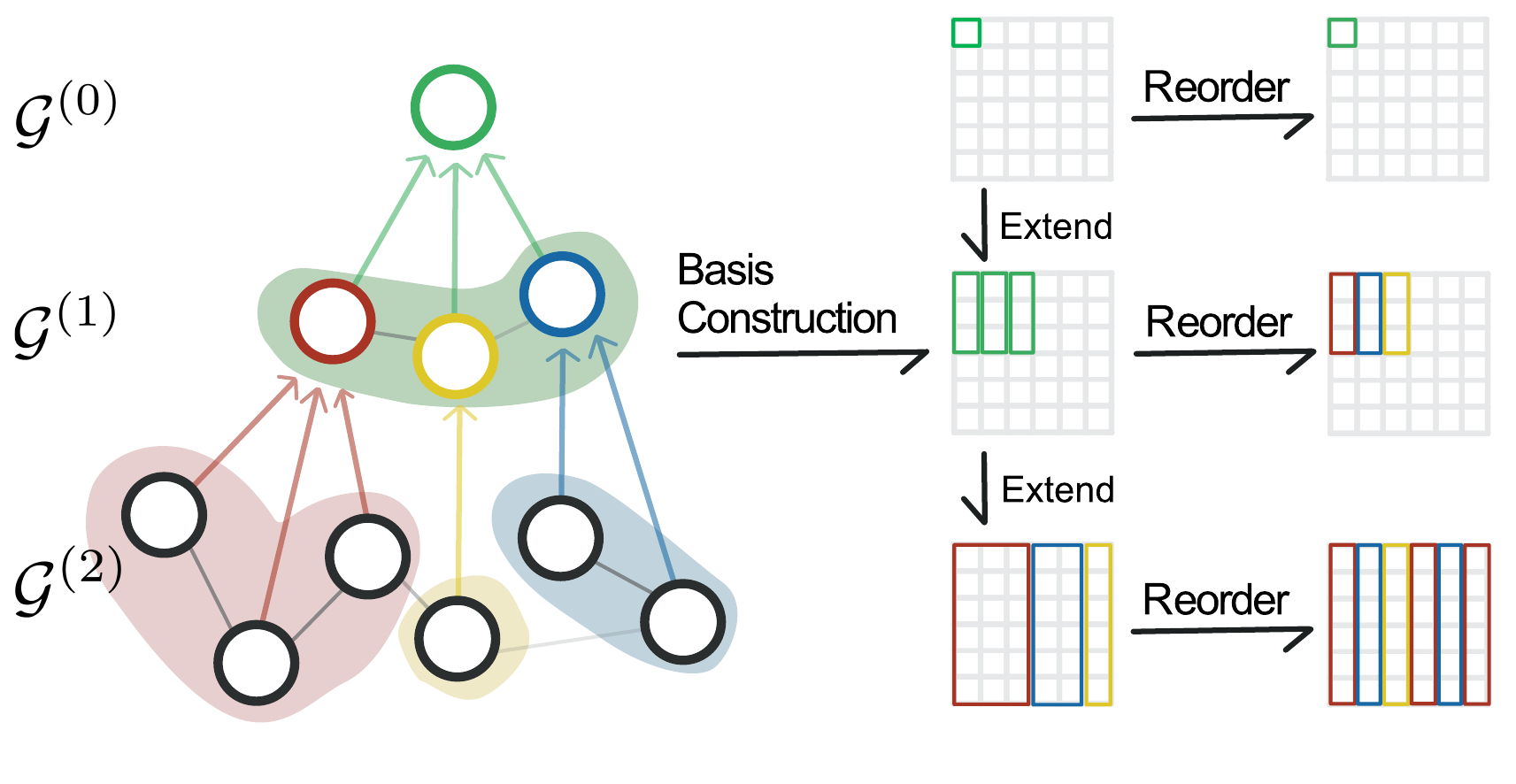}
    \caption{For a given coarse-grained chain $\mathcal{G}^{J \rightarrow J_0}$, MRA constructs a global orthogonal basis in a top-down fashion, see Algorithm~\ref{algo3-GOB}. At level $j$, each parent basis is dilated to level $j+1$, where the number of dilated bases are determined by its children number at level $j+1$. The generated basis is sparse and properly sorted.}
    \label{fig2}
  \end{minipage}
\end{figure}

The fast transform on wavelets with orthonormal bases relies on a hierarchical architecture, which can be characterized by a coarse-grained chain algorithm. The complete chain structure can be designed via either spectral or non-spectral clustering algorithms. Given the finest level of a single graph $\mathcal{G}^{(J)}\coloneqq\mathcal{G}$, we group $N^{(J)}$ vertices into $N^{(J-1)}$ clusters. The newly constructed $\mathcal{G}^{(J-1)}$ is called a coarse-grained graph of $\mathcal{G}^{(J)}$, where each vertex $[v]\in \mathcal{V}^{(J-1)}$ also represents a cluster of vertices $v \in \mathcal{V}^{(J)}$ in level $J$. Each of the new vertices $[v]$ in $\mathcal{V}^{(J-1)}$ is picked from the corresponding cluster as their `representatives', and the new weighted adjacency matrix $\mathcal{W}^{(J-1)}$ is obtained accordingly. For two vertices $[p],[v]\in \mathcal{V}^{(J-1)}$, their pairwise weight is
\begin{equation}
    \mathcal{W}^{(J-1)}([p],[v])\coloneqq\sum_p\sum_v\frac{\mathcal{W}^{(J)}([p],[v])}{\text{vol}(\mathcal{G}^{(J)})}. \label{weightAdj}
\end{equation}
where $\text{vol}(\mathcal{G}^{(J)}) \coloneqq \text{vol}(\mathcal{V}) = \sum_v \sum_p \mathcal{W}^{(J)}(p,v)$ is the volume of the graph $\mathcal{G}^{(J)}$, i.e., the sum of degrees of all vertices in $\mathcal{G}^{(J)}$. The summation in \Eqref{weightAdj} is taken along all pairs of $p\in[p]$ and $v\in[v]$. Once $\mathcal{W}^{(J-1)}$ is fully constructed, we can repeat the same procedure to build the adjacency matrix $\{\mathcal{W}^{(J-2)},\dots,\mathcal{W}^{(J_0)}\}$ on the coarse-grained graphs until $\mathcal{W}^{(J_0)}$ of the coarsest level $\mathcal{G}^{(J_0)}$. The generated sequence of subspaces naturally resembles a multi resolutions of $\mathcal{G}$, which promises developing graph wavelet bases. 

There are many clustering methods, such as $k$-means \cite{lloyd1982least}, METIS \cite{karypis1995metis} and spectral clustering \cite{von2007tutorial}. By the time complexity, the METIS is faster than spectral clustering and $k$-means. In terms of the clustering performance, spectral clustering and $k$-means are more robust than METIS. In this work we will consider the spectral clustering method.

\subsection{Orthonormal Basis Construction}
When the tree representation is prepared, one can  extend a Haar-like orthonormal wavelet basis on $\mathcal{G}$. The construction of global orthonormal basis has been studied extensively in the literature \cite{li2019haar,chui2015representation,gavish2012sampling,wang2019haarpooling,wang2019tight}. Constructing an orthonormal Haar-like Wavelet-based MRA system generally requires extending parent basis functions progressively from the coarsest level to the finest level of the chain, see Algorithm~\ref{algo3-GOB}. In addition, a \textit{characteristic function} is required for defining the mother wavelet on non-Euclidean graph data. We present the mathematics as below.

Recall in Section~\ref{mra} we define a wavelet-like orthonormal basis of MRA as
\begin{equation*}
    \Phi = \{\Phi^{(J_0)}, \Psi^{(J_0)}, \dots, \Psi^{(J)}\}.
    \label{partialGOB}
\end{equation*}
The above MRA system is a key to fast decomposition and reconstruction. In the context of graph representation, we prepare an orthonormal basis to every level of the chain so that they are comparable to each coarse-grained graph. We can apply \Eqref{projection} to find the full set of basis at level $j$ by $\Phi^{(j)}=\Phi^{(j-1)*}+\Psi^{(j-1)}$ with $\Phi^{(j-1)*}$ representing vertically extended $\Phi^{(j-1)}$. This gives a sequence of orthonormal basis (or a scaling-like system)
\begin{equation}
\Phi = \{\Phi^{(J_0)}, \Phi^{(J_0+1)}, \dots, \Phi^{(J)}\},
\label{GOB}
\end{equation} 
where $\Phi^{(j)}$ is the basis matrix corresponding to the graph $\mathcal{G}^{(j)} = (\mathcal{V}^{(j)}, \mathcal{E}^{(j)}, \mathcal{W}^{(j)})$ at the $j$th level of the chain. The dimension of each basis matrix is determined by the number of vertices at that level, i.e., $\Phi^{(j)} = \bigl(\phi_1^{(j)}, \dots, \phi_{N^{(j)}}^{(j)}\bigr)$ is composed of $N^{(j)}\coloneqq |\mathcal{V}_j|$ basis vectors, which is equal to the number of vertices of the $j$-level coarse-grained graph. As a result, constructing $\Phi^{(j)}$ from $\Phi^{(j-1)}$ requires the size of the orthogonal system to be extended from $N^{(j-1)} \times N^{(j-1)}$ to $N^{(j)} \times N^{(j)}$. In particular, we call $\Phi^{(J)}$ at the finest level $\mathcal{G}^{(J)}$ a \emph{global orthonormal basis} of the graph $\mathcal{G}$. A tree-structured global basis is useful for wavelet decomposition and reconstruction, during which the pair-wise interaction in the input signal is sparsified and the mutual information between local and global embeddings is reserved \cite{ye2018deep}.

\subsubsection{Characteristic Function}
As far as the general idea has already be introduced, constructing graph-based wavelets could still be tricky since a parent vertex could have an arbitrary number of children in the next finer level. Also the non-Euclidean data structure makes it less obvious to define `neighbors' or processing windows as in conventional signal processing procedures. Here we consider the simple dyadic cases that describe `parent-children' relationships by a piece-wise characteristic function. By `dyadic' we need each vertex to only have one parent. We also require that a parent has at least two children node. This is a critical property as it guarantee the Haar transforms to have linear computational complexity. (See []) We call a tree with this piece-wise dyadic splitting manner a Haar-like filtration, where the tree describes the `parent-children' relationships of vertices. Here we follow \cite{chui2015representation} and define the characteristic function $\chi_{k}^{(j)},1\leq k\leq N^{(j-1)}$ for the $k$th cluster in $\mathcal{V}^{(j)}$,
\begin{equation}
    \chi_{k}^{(j)}:=\left\{
\begin{array}{ll}
1, & [v_{\ell}^{(j)}]=v_{k}^{(j-1)}, \\
0, & \text{otherwise}. 
\end{array}\right.
\end{equation}
The above function determine whether the $\ell$th vertex $v_{\ell}^{(j)},1\leq \ell \leq N^{(j)}$ belongs to the $k$th cluster of $\mathcal{V}^{(j)}$. The above process is associated with the $k$th vertex $v_{k}^{(j-1)}\in\mathcal{V}^{(j-1)}$, and the condition of $\chi_{k}^{(j)} = 1$ is only met when $v_{k}^{(j-1)}$ is the parent of $v_{\ell}^{(j)}$ at level $j-1$. This dyadic filtration ensures a sparse establishment of $\phi^{(j)}_{\ell,k}(v)$, where the basis $\phi^{(j-1)}_k([v])$ with respect to $k$ only supports on its children, and only the members of level $j$ in $[v_{\ell,k}^{(j)}]$ will have nonzero basis values. 

\subsubsection{Extension Function}
Now that we have an explicit rule for the `parent-children' relationship, the next step is to establish the rule to pass information from each parent to its children. Instead of handling the full set of column vectors $\{\phi_{1}^{(j-1)}([v]), ..., \phi_{N^{(j-1)}}^{(j-1)}([v])\}$, we start from the $k$th basis vector $\phi^{(j-1)}_k([v]) \in \mathbb{R}^{N^{(j-1)}}$ with a filtration $\chi_{k}^{(j)}$. The target is to extend this parent basis to its $n_{k}^{(j)}$ children and obtain a sequence of basis vectors $\{\phi_{1,k}^{(j)}(v), ..., \phi_{n_{k}^{(j)},k}^{(j)}(v)\}, v\in[v]$ at level $j$. The $n_{k}^{(j)}$ is the size of the $k$th cluster at the $j$th level. We will divide the process into two steps as the vertical and the horizontal extensions. %recursively along the coarse-grained chain $\mathcal{G}^{J \rightarrow J_0}$. 

We start from the vertical extension and obtain the $\phi_{k}^{(j)*}([v])$ of length $N^{(j)}$ by
\begin{equation}
    \phi_{k}^{(j)*}([v]) \coloneqq     \frac{\phi_{k}^{(j-1)}([v])}{\sqrt{n_{k}^{(j)}}}.
    \label{eq:basisExtendV}
\end{equation}
The vertices $v\in[v]$ share the same parent $v_{k}^{(j-1)} \in \mathcal{V}^{(j-1)}$ that is from the coarser level $j-1$.

The next step is to extend the parent $\phi_{k}^{(j)*}([v])$ horizontally and obtain the rest $n_{k}^{(j)}-1$ vectors $\{\phi_{2,k}^{(j)}(v), ..., \phi_{n_{k}^{(j)},k}^{(j)}(v)\}$. 
% This can be practiced by many techniques. For example, with Gram-Schmidt process \cite{arfken1999mathematical}, the $\ell$th new basis of the $k$th cluster is
% \begin{equation*}
%     \phi_{\ell,k}^{(j)}(v) \coloneqq \phi_{k}^{(j)*}([v]) - \sum_{i=1}^{\ell-1} \operatorname{Proj}_{\mathcal{G}^{(j)}} (\phi_{k}^{(j)*}([v])).
% \end{equation*}
The final horizontal extension function of the system also includes the normalization factor and the characteristic function. For the system $\{\phi^{(j-1)}_k,\chi_{k,1},\dots,\chi_{k,n_{k}^{(j)}}\}$, we define the $\ell$th basis
\begin{equation}
\phi_{\ell,k}^{(j)}(v) \coloneqq \sqrt{\frac{n_{k}^{(j)}-\ell+1}{n_{k}^{(j)}-\ell+2}}\left(\chi_{\ell-1,k}^{(j)}-\frac{\sum_{i=\ell}^{n_{k}^{(j)}} \chi_{i,k}^{(j)}}{n_{k}^{(j)}-\ell+1}\right), \quad \ell=1,\dots,n_{k}^{(j)}.
\label{basisExtendH}
\end{equation}

The same extension function can be applied to the other $N^{(j-1)}-1$ basis in level $j-1$ to finish the process. For the $\Phi^{(j-1)} \in \R^{N^{(j-1)}\times N^{(j-1)}}$, one first applies \Eqref{eq:basisExtendV} and obtains $\Phi^{(j-1)*} \in \R^{N^{(j-1)}\times N^{(j)}}$. During this step, the vectors are stretched to a longer form, but no new vector is created. Then in \Eqref{basisExtendH}, the vertically extended basis vectors are dilated within the filtration, i.e., each parent only passes the information to its children. As a result, a new orthonormal basis $\Phi^{(j)} \in \R^{N^{(j)}\times N^{(j)}}$ is sorted for $L_2(\mathcal{G}^{(j+1)})$ at the next $j$ level. This extension scheme is universally applicable on any level of the chain $\mathcal{G}^{J \rightarrow J_0+1}$. We call $\Phi^{(j)}$ a \emph{Haar-like basis} on $L_2(\mathcal{G}^{(j)})$. Specifically, when $j = J$, the corresponding $\Phi^{(J)}$ at the finest level is called the \textit{Global Haar-like Orthonormal Basis} of graph $\mathcal{G}$.

One special case is at the coarsest level $J_0$, where each vertex represents an exclusive cluster, and there is no further coarser level `$J_{-1}$'. We modify \Eqref{eq:basisExtendV} and \Eqref{basisExtendH} accordingly by defining $n_{\ell}^{(J_0)} = N^{(J_0)}$. For $\ell = 2, 3, ..., N^{(J_0)}$, we then have
\begin{equation}
\begin{aligned}
\phi_{1}^{(J_0)} &\coloneqq \frac{1}{\sqrt{N^{(J_0)}}};\\[1mm]
\phi_{\ell}^{(J_0)} &\coloneqq \sqrt{\frac{N^{(J_0)}-\ell+1}{N^{(J_0)}-\ell+2}}\left(\chi_{\ell-1}^{(J_0)}-\frac{\sum_{i=\ell}^{N^{(J_0)}} \chi_{i}^{(J_0)}}{N^{(J_0)}-\ell+1}\right). \notag
\end{aligned}
\end{equation}
Note that we define a uniform function of $\phi_{\ell}^{(j-1)}([v]) = 1$ in the coarsest level $J_0$ for simplification. However, this function could be replaced by any other standardized functions.

The general recipe of constructing a Global Haar-like Orthonormal Basis is summarized in Algorithm~\ref{algo3-GOB}. For a better understanding, this process is also visualized in Figure~\ref{fig2}. In the next two sections, we demonstrate how the prepared Haar-like orthonormal system $\{\phi_{\ell}^{(j)}\}_{\ell=1}^{N^{(j)}}$, $j=J_0, \dots, J$ supports the graph convolution and graph pooling operators.

\begin{algorithm}[t]
\SetKwData{step}{Step}
\SetKwInOut{Input}{Input}\SetKwInOut{Output}{Output}
\BlankLine
\Input{A coarse-grained chain $\mathcal{G}^{J \rightarrow J_0}$ with $\mathcal{G}^{(j)} = (\mathcal{V}^{(j)}, \mathcal{E}^{(j)}, \mathcal{W}^{(j)})$.}
\Output{A global orthonormal basis $\Phi$ of graph $\mathcal{G}$.}
Initialization: $j=J_0+1$.\\%, n_{\ell}^{(J_0)} = N^{(J_0)}, N^{(j-1)}=|\mathcal{V}^{(J_0)}|$.\\
\While{$j\leq J$}
{
    From $\Phi^{(j-1)} \in \mathbb{R}^{N^{(j-1)}\times N^{(j-1)}}$ construct basis vectors $\Phi^{(j)*} \in \mathbb{R}^{N^{(j)}\times N^{(j-1)}}$ by \Eqref{eq:basisExtendV}.\\    \tcp*[f]{Vertical Extension}\\
    
    Extend $\Phi^{(j)*} \in \mathbb{R}^{N^{(j)}\times N^{(j-1)}}$ to $\{\phi_{\ell}^{(j)}\}_{\ell=1}^{N^{(j)}} \in \mathbb{R}^{N^{(j)}\times N^{(j)}}$ by calculating the rest $N^{(j)}-N^{(j-1)}$ basis vectors from \Eqref{basisExtendH}. \tcp*[f]{Horizontal Extension}\\
    
    Rearrange $\{\phi_{\ell}^{(j)}\}$ to an orthonormal basis $\Phi^{(j)}$ on $\mathcal{G}^{(j)}$ so that the first $N^{(j-1)}$ vectors are orthonormal to $L_2(\mathcal{G}^{(j-1)})$. \tcp*[f]{Basis Update}\\
}
\caption{Haar-like-based MRA Global Orthonormal Basis}
\label{algo3-GOB}
%\vspace{-0.6cm}
\end{algorithm}

\subsection{Graph Convolution}
The classic spectral graph convolution \cite{bruna2013spectral} relies on the Fourier basis of the graph Laplacian. We follow the convention and define the spectral graph convolution for an arbitrary graph, which transforms the input signal $X^{\text{in}}$ with an orthonormal basis and a trainable diagonal filter $G$
\begin{align}
 \widetilde{X}^{\text{in}} &= X^{\text{in}}W;\notag\\
    X^{\text{out}} &= \sigma(\Phi G \Phi^{-1} \widetilde{X}^{\text{in}}),\label{spectralConv}
\end{align}
where $\sigma(\cdot)$ is the non-linear activation function. The weight matrix $W\in \mathbb{R}^{d^{\rm in}\times d^{\rm out}}$ provides an affine transform of the input $X^{\rm in}\in \mathbb{R}^{N\times d^{\rm in}}$ to $\widetilde{X}^{\rm in}\in \mathbb{R}^{N\times d^{\rm out}}$. Consider this convolution process as (discrete) forward and adjoint wavelet transforms, then the forward Fourier transform $\Phi^{-1}$ maps $\widetilde{X}^{\rm in}$ from the vertex domain to the Fourier domain; the diagonal filter $G$ processes feature information with the transformed graph data; and the adjoint wavelet transform $\Phi$ projects the information back to the vertex domain to accomplish the convolution. The network trainable filter $G$ controls the information processing in GNNs by determining which part of the signals should be preserved, passed, or filtered out. Note that \Eqref{spectralConv} includes a weight detaching trick, that is, conducting feature transformation $X^{\text{in}}W$ before performing the graph convolution, to reduce the parameter complexity from $\mathcal{O}(N\times d^{\rm in}\times d^{\rm out})$ to $\mathcal{O}(N+d^{\rm in}\times d^{\rm out})$ \cite{xu2018graph, li2019haar}.

Although theoretically sound, the computational cost of eigendecomposition and discrete Fourier transform in \Eqref{spectralConv} increases drastically with the graph size, which in no circumstance is desirable. Instead, in this work, we rule out the Fourier basis with a Haar-like orthonormal basis system from MRA to avoid a dense matrix's fussy computation. The coarse-grained chain structure of graph data supports a compressive workflow of generating a set of the sparse and localized operator $\Phi$, enabling a fast implementation of the forward and adjoint Haar transforms for graph signals between the vertex domain and wavelet domain. This fast computation algorithms then promise efficient graph convolution and pooling operations, as we introduce now. %To support the empirical network training procedure, we propose a padding strategy for the global structure of filtration.

%Many approaches so far have been proposed to generate spectral-based convolution. For example, graph Fourier transforms \cite{shuman2013emerging} define the transform matrix $\Phi$ by the graph Laplacian, and graph wavelet neural networks \cite{xu2018graph} use the graph wavelet transforms, which provides a sparse and localized convolution for GNN.
\subsubsection{Fast Forward Haar-like Wavelet Transform}
In this section, we present a fast algorithm for graph convolution. For the sake of simplicity, we consider input $x\in \mathbb{R}^{N}$ with one feature. One can easily generalize the calculation to higher-dimension features. We start from Fast Forward Haar-like Wavelet Transform that processes the raw $x$ from the vertex domain to the wavelet domain. At the $j$th level, we expect to find a set of Haar-like wavelet coefficients
$$
\bigl(c_{1}^{(j)}, ..., c_{N^{(j)}}^{(j)}\bigr) \in \mathbb{R}^{N^{(j)}}.
$$
through $\Phi^{\top} x\coloneqq \langle x, \phi_{\ell}^{(J)} \rangle$, $N^{(j-1)}<\ell\leq N^{(j)}$. In this set of $N^{(j)}$ coefficients, the first $N^{(j-1)}$ coefficients are inherited from the coarser $(j-1)$-level. The remaining $N^{(j)}-N^{(j-1)}$ coefficients are from the `newly constructed' basis vectors in this level, which we call the detail coefficients. So instead of calculating the full inner product at the finest level $J$, we can only calculate the detail coefficients at each level. It avoids repeated computation on the approximation coefficients that later become detail coefficients in some coarser level.

For $j=J_0,\dots,J$, suppose $N^{(j-1)}<\ell\leq N^{(j)}$. We write the $\ell$th coefficient for graph signal $x$ in term of the basis expansion of the $j$th level by
%At level $j-1$, the coefficient is assembled by vertices information from a finer level $j$. Given $N^{(j-1)}<\ell\leq N^{(j)}, v\in[v]$ in $\mathcal{V}^{(j)}$ and $[v]\in\mathcal{V}^{(j-1)}$ for some $J_0\leq j\leq J$, the $\ell$th coefficient is
\begin{align*}
\langle x, \phi_{\ell} \rangle 
&= \sum_{v\in \mathcal{V}^{(J)}} x(v) \phi_{\ell}^{(J)}(v)
= \sum_{[v]_{\gph^{(j)}}\in \mathcal{V}^{(j)}}\sum_{u\in [v]_{\gph^{(j)}}} x(u) \phi_{\ell}^{(j)}(u)\\
&= \sum_{[v]_{}\in \mathcal{V}^{(j)}}\phi_{\ell}^{(j)*}([v]) \sum_{u\in[v]_{\gph^{(j)}}}x(u)
= \sum_{[v]\in \mathcal{V}^{(j)}}\phi_{\ell}^{(j)*}([v])S_{[v]}^{(j)}(x).
\end{align*}
The $v$ here, rather than an arbitrary vertex of $\mathcal{G}$, is one of those parent vertex at level $J$ that becomes a child vertex at level $j$. We define $S_{[v]}^{(j)}(x) \coloneqq \sum_{u\in[v]_{\gph^{(j)}}}x(u)$ to describe the `super instance' $x([v])$ with respect to the cluster $[v]=[v]_{\gph^{(j)}}$ at the $j$-th level. This summation contains the information aggregated from all children instances of $[v]$ from the finer $J$-level.

Now we include the normalization from \Eqref{eq:basisExtendV}, and wrap up the formal procedure for calculating the wavelet coefficients, that is, the fast forward Haar-like wavelet transform.
\begin{align}
c_{\ell}^{(j)} &\coloneqq 
\sum_{[v]\in \mathcal{V}^{(j)}}\phi_{\ell}^{(j)}([v])\frac{S_{[v]}^{(j)}(x)}{\sqrt{|[v]|}}, \label{forwardConv}
 \end{align}
and for $N^{(j-1)}<\ell\leq N^{(j)}$, we can calculate the coefficient recursively by
\begin{align}
S_{[v]}^{(j)}(x) &\coloneqq \sum_{[u]_{\gph^{(j+1)}}\in [v]_{\gph^{(j)}}} \frac{S_{[u]}^{(j+1)}(x)}{\sqrt{|[u]|}}. \label{weightedSum}
\end{align}
% For every $\ell$, the sum product $S_{\ell}^{(j)}(x)$ is normalized by a weight factor $\frac1{\sqrt{n_{\ell}^{(j)}}}$. 
The weighted sum information of all the children vertices $[u]_{\gph^{(j+1)}}$ is used to describe the parent vertex $[v]_{\gph^{(j)}}$ of the coarser level $j$. The $\ell$th coefficient $c_{\ell}^{(j)}$ is calculated by the definition of inner product, as is in \Eqref{forwardConv} when the sum $S_{[v]}^{(j)}(x)$ is evaluated (recurrently). The full set of wavelet coefficients is fast computed by this bottom-up manner. We summarize the pseudocode of our Fast Forward Haar-like Wavelet Transform in Algorithm~\ref{algo1-FDWT}. 

\begin{algorithm}[t]
\SetKwData{step}{Step}
\SetKwInOut{Input}{Input}\SetKwInOut{Output}{Output}
\BlankLine
\Input{A global orthonormal basis $\Phi=(\phi_{\ell})_{\ell=1}^{N^{(J)}}$; a vector $x$.}
\Output{Wavelet coefficients $\mathbf{c} = (c_{\ell})_{\ell=1}^{N^{(J)}}$.}
Initialization: $N^{(J)} = |\mathcal{V}|;\; S_v^{(J+1)}(x)=x(v), \; v \in \mathcal{V}$.\\
\For{$j$=$J$ \KwTo $J_0$}
{
    Prepare $S_{\ell}^{(j)}(x)$ by \Eqref{weightedSum}. \tcp*[f]{Weighted Summation}\\
    \For{$\ell$=$N^{(j-1)}+1$ \KwTo $N^{(j)}$}
    {
        Project $x$ to $c_{\ell}^{(j)}$ by \Eqref{forwardConv}. \tcp*[f]{Coefficients Calculation}
    }
}
\caption{Fast Forward Haar-like Wavelet Transform on $\mathcal{G}$}% (FDWT)}
\label{algo1-FDWT}
%\vspace{-0.6cm}
\end{algorithm}

\begin{algorithm}[t]
\SetKwData{step}{Step}
\SetKwInOut{Input}{Input}\SetKwInOut{Output}{Output}
\BlankLine
\Input{A global orthonormal basis $\Phi=(\phi_{\ell})_{\ell=1}^{N^{(J)}}$; wavelet coefficients $\mathbf{c} = (c_{\ell})_{\ell=1}^{N^{(J)}}$.}
\Output{vector $x_J$.}
Initialization: $N^{(J_0-1)} = 0$; $x(v) = 0\; \forall v\in\mathcal{G}^{(J_0)}$.\\
\For{$j=J_0$ \KwTo $J$}
{
    %$N^{(j)} = |\mathcal{V}^{(j)}|$.\\
    $\tv{[v]}(\mathbf{c}) = 0$ for all $[v] \in \mathcal{V}_j$.\\
    \For{$\ell=N^{(j-1)}+1$ \KwTo $N^{(j)}$}
    {
        Find $\tv{[v]}(\mathbf{c})$ and the weight $W_{k}^{(j)}$ by \Eqref{weightReconstruction} for all $[v] \in \mathcal{V}_j$.\\% = S_{v}^{(j-1)}(c)+c_{\ell}\phi_{\ell}^{(j)*}([v])$.
        \tcp*[f]{Weighted Summation}
    }
    Update $x_J(v)$ by \Eqref{reconstruction} for all $v \in \mathcal{V}_J$. \tcp*[f]{Signal Reconstruction}
}
\caption{Fast Adjoint Haar-like Wavelet Transform on $\mathcal{G}$}% (FDWT)}
\label{algo2-FADWT}
%\vspace{-0.6cm}
\end{algorithm}

\subsubsection{Fast Adjoint Haar-like Wavelet Transform}
Given a sequence of wavelet coefficients, the adjoint Haar-like transform can reconstruct the signal from the coefficients. In this section we introduce the fast implementation of the adjoint transform. For $N^{(j-1)}<\ell\leq N^{(j)}$ and $j=J_0,\dots,J$, the raw $x$ is reconstructed through
\begin{align*}
x(v)
&= \sum_{\ell=1}^{N^{(j)}} c_{\ell}\phi_{\ell}(v)
= \sum_{j=J_0}^{J} \sum_{\ell=N^{(j-1)}+1}^{N^{(j)}} c_{\ell}\phi_{\ell}(v)
= \sum_{j=J_0}^{J} \sum_{\ell=N^{(j-1)}+1}^{N^{(j)}}\sum_{[v]\in \mathcal{V}^{(j)}}c_{\ell}\phi_{\ell}^{(j)}([v])\cdot \mathbf{1}_{[v]}\\
&= \sum_{j=J_0}^{J} \sum_{[v]\in \mathcal{V}^{(j)}} \left(\sum_{\ell=N^{(j-1)}+1}^{N^{(j)}}c_{\ell}\phi_{\ell}^{(j)}([v])\right)\mathbf{1}_{v}
= \sum_{j=J_0}^{J} \sum_{[v]\in \mathcal{V}^{(j)}} \tv{[v]}(\mathbf{c})\mathbf{1}_{[v]}
\end{align*}
where $\mathbf{c}=(c_{\ell})_{\ell=1}^{N^{(j)}}$ is a sequence of coefficients on $\mathcal{G}^{(j)}$ and $\tv{[v]}(\mathbf{c}) \coloneqq \sum_{\ell=1}^{N^{(j)}}c_{\ell}\phi_{\ell}^{(j)}([v])$ with $v \in \mathcal{V}^{(j)}$. The $\mathbf{1}_{[v]}\coloneqq \mathcal{V}\to \mathbb{R}$ is an indicator function by $\mathbf{1}_{[v]}(v)=1$ for $v\in[v]$. 
% The above process is connected to \Eqref{projection} in the sense that $\operatorname{Proj}_{\widetilde{\mathcal{G}^{(j)}}} \mathbf{c} = \sum_{[v]\in \mathcal{V}^{(j)}} \tv{[v]}(\mathbf{c})\mathbf{1}_{[v]}$. 
Here, in order to reconstruct the full $x$, we start from the top level of the hierarchical chain and find the approximation information $x^{(J_0)}$ with respect to the nodes. For the $j$th level finer than the $J_0$th, only $N^{(j)}-N^{(j-1)}$ detail coefficients are transformed with the corresponding basis $\phi^{(j)}([v])$ and then aggregated as $x^{(j)}$.

In correspondence to the forward process of \Eqref{forwardConv}, since the coefficients are normalized, the adjoint transform would also require a weight matrix on the accumulated coefficients $\mathbf{c}$ when restoring the original scale of the input data. Given a set of coefficients $\mathbf{c} = \{c_{1}, ..., c_{N}\}$, we compute the reconstructed graph signal for $v\in\gph$ by
\begin{align}
%\Phi c_{\ell} 
x_J(v) &\coloneqq \sum_{j=J_0}^{J} W^{(j)} \tv{[v]}(\mathbf{c}),
\label{reconstruction}
\end{align}
where for $j=J_{0},\dots,J-1$,
\begin{align}
\tv{[v]}(\mathbf{c}) \coloneqq \sum_{\ell = N^{(j-1)}+1}^{N^{(j)}} c_{\ell} \phi_{\ell}^{(j)*}([v]),\quad
W^{(j)} \coloneqq \prod_{\ell=2}^{j} \sqrt{n_{\ell}^{(j)}},
\label{weightReconstruction}
\end{align}
where $[v]$ in $\phi_{\ell}^{(j)*}([v])$ is the parent of $v\in\mathcal{G}$ at the $j$th level. The formulation shows that the vertex information of an arbitrary $v\in\mathcal{G}^{(J)}$ is reconstructed by collecting the weighted information from all its parents $[v]\in\mathcal{G}^{(j)},J_0\leq j\leq J-1$. We summarize in Algorithm~\ref{algo2-FADWT} for this top-down computational method. This procedure, in total, requires a maximum number of summations of $\sum_{j=J_0}^{J-1} N^{(j+1)}$.

%\begin{minipage}{0.47\textwidth}
%algo1
%\end{minipage}
%\hfill
%\begin{minipage}{0.47\textwidth}
%algo2
%\end{minipage}
We shall emphasize again that the above computation is applied to a single graph. In graph property prediction tasks, each graph will be transformed individually. However, they do share a same filter for data processing. This trainable filter matrix, denoted by $G$, is learned in the wavelet domain and it is used to process the Haar coefficients. Conventionally, the filter matrix for an individual graph $\mathcal{G}^{(j)}$ is constructed as a diagonal matrix of size $N^{(j)}$. In graph-level tasks, different graphs share the same number of features, but contain varying number of nodes. To eliminate the inconsistency and apply an identical filtration rule along the graph data set, in practice, we consider a padding strategy on $\Phi$ and then extend all bases to a unified dimension. We fill the extra columns by $\mathbf{0}\in\mathbb{R}^{N^{(j)}}$ vectors. The shape of the modified basis matrix is then $\Phi^{(j)} \in \mathbb{R}^{N^{(j)} \times N_{\text{max}}^{(j)}}$, $N_{\text{max}}^{(j)} \coloneqq \max\{N_1^{(j)}, ..., N_M^{(j)}\}$.

\subsection{Graph Pooling}
For graph properties prediction tasks, a pooling operation is usually required to unify graph embedding for a given set of graphs with different structures and sizes. Instead of preprocessing the graph data to a consistent size, our model relies on the constructed orthonormal basis and performs \emph{Haar-like Wavelet Pooling} along the coarse-grained chain. For instance, an input graph $\mathcal{G}$ with a chain $\mathcal{G}^{J\to J_0}$ requires $J-J_0$ wavelet pooling layers in GNN to reach the final graph embedding. In particular, we define the wavelet pooling of the $j$th layer by
\begin{equation}
\widehat{X}^{(j)} = (\widetilde{\Phi}^{(j)})^T X^{(j)} 
\end{equation}
for $j= J, J-1,\dots, J_0+1$, where $X^{(j)} \in \mathbb{R}^{N^{(j)} \times d}$ is the output from the previous convolutional layer. The pooling operator $\widetilde{\Phi}^{(j)}$ takes the first $N^{(j-1)}$ columns from the associated orthonormal basis $\Phi^{(j)}$ at level $j$ of the chain. As a result, the output of this Haar-like Wavelet Pooling layer downsamples the matrix size to $\mathbb{R}^{N^{(j-1)} \times d}$ for all the graphs. The information of graphs is then compressed progressively until the last layer, where all the graphs have a standard output size of $\mathbb{R}^{1\times d'}$. The output feature dimension $d'$ is determined by the learnable detaching matrix $W$ in the corresponding graph convolution layer.

% Given that the basis system $\Phi$ is built with MRA, we can adopt the compressive wavelet transform so that the pooling strategy possesses fast implementation. To minimize the number of summation and multiplication operations, we modify the weighted sum of the signal information in \Eqref{weightedSum} by pre-gathering the feature vectors from child nodes to their coarsest possible level. Formally, the sum-product for an input feature $x^{(j)} \in \mathbb{R}^{N^{(j)}\times d}$ is
% \begin{align}
% S^{(J+1)}(x) &\coloneqq x^{(j)} \label{parent} \\
% S_{v_{\ell}}^{(i)}(x) &\coloneqq \sum_{v_{\ell'}^{(i+1)} \in v_{\ell}^{(i)}} \frac{S_{v_{\ell'}}^{(i+1)}}{\sqrt{n_{\ell'}^{(i+1)}}}. \label{children}
% \end{align}
% The $\phi_{\ell}^{(i)}$ is the $\ell$th member of the orthonormal basis $\{\phi_{\ell}^{(i)}\}_{\ell=1}^{N^{(i)}}$, where $i \leq j$ represents the coarsest possible layer of the $\ell$th vertex. At each level $j$, the compressive Haar-like transform in \Eqref{parent} inherits full information of the first $N^{(j-1)}$ feature vectors which are elected as parent nodes at level $j-1$. The rest $N^{(j)} - N^{(j-1)}$ child nodes contribute their feature information by \Eqref{children} to the coarsest possible layer. Again, the sum product includes a weight factor $\frac1{\sqrt{n_{\ell}^{(j)}}}$ to balance the information importance from clusters of difference sizes. Given the sum product $S^{(j)}(x)$ from parents and $S_{\ell}^{(i)}(x)$ from children, one can then send them to \Eqref{forwardConv} and finish the compressive Haar-like transform.

To understand the rationality of adopting this pooling strategy, we revisit \Eqref{projection} in Section~\ref{mra}. Recall that the whole decomposition is guided by the pyramid algorithm, where each coarse-grained graph $\mathcal{G}^{(j)}$ promises its first $N^{(j-1)}$ low-frequency coefficients reflect the approximation of the original signal $X$. The rest $N^{(j)}-N^{(j-1)}$ detail wavelet coefficients preserve complementary information, which are essentially noise terms. Through the removal of high-frequency components, this approach of Wavelet Compression is capable of reducing a graph's size while simultaneously preserving its key information. Furthermore, this compressive pooling strategy combines neighborhood information similarly as in the average pooling. With the first $N^{(j-1)}$ basis vectors of $\Phi^{(j)}$ from \Eqref{eq:basisExtendV}, we calculate the wavelet coefficient for the $\ell$th ($1\leq\ell\leq N^{(j-1)}$) basis vector at the $j$th level by
\begin{align*}
\langle x^{(j)}, \phi_{\ell}^{(j)} \rangle 
&= \sum_{v\in \mathcal{V}^{(j)}} x^{(j)}(v) \phi_{\ell}^{(j)}(v)
= \sum_{[u]\in \mathcal{V}^{(j-1)}} \sum_{v \in [u]} x^{(j)}([u]) \frac{\phi_{\ell}^{(j-1)}([u])}{\sqrt{|[u]|}}\\
&= \sum_{[u]\in \mathcal{V}^{(j-1)}} \left(\frac{\sum_{v \in [u]} x^{(j)}(v)}{\sqrt{|[u]|}}\right) \phi_{\ell}^{(j-1)}([u])
= \langle \widetilde{x}^{(j)},\phi^{(j-1)}_{\ell} \rangle 
\end{align*}
where $v\in\mathcal{V}^{(j)}$ belongs to one of the clusters $[u]=[u]_{\gph^{(j-1)}}\in \mathcal{V}^{(j-1)}$ at the coarser level $j-1$. The $\widetilde{x}^{(j)}([u]) \coloneqq  \frac{\sum_{v \in [u]} x^{(j)}(v)}{\sqrt{|[u]|}}$ is a weighted `super instance' with $[u]$ from $x^{(j)}(v)$ of its children, and $|[u]|$ is the number of nodes from the $j$th level which belong to the cluster $[u]_{\gph^{(j-1)}}$. The above derivation implies that
$$
\sum_{\ell=1}^{N^{(j)}}\bigl|\langle x^{(j)}, \widetilde{\Phi}_{\ell}^{(j)}\rangle \bigr|^2 = \sum_{\ell=1}^{N^{(j)}}\bigl|\langle \widetilde{x}^{(j)}, \Phi_{\ell}^{(j-1)} \rangle \bigr|^2.
$$
For an orthonormal basis $\Phi^{(j-1)}$ of $L^2(\mathcal{G}^{(j-1)})$, we can then deduce the relationship
\begin{align*}
\bigl\lVert (\widetilde{\Phi}^{(j)})^{\top} x^{(j)}\bigr\rVert^2 
= \sum_{\ell=1}^{N^{(j)}}\bigl|\langle \widetilde{x}^{(j)}, \Phi_{\ell}^{(j-1)} \rangle\bigr|^2 
= \bigl\lVert \widetilde{x}^{(j)}\bigr\rVert^2
= \sum_{[u]\in \mathcal{V}^{(j-1)}} \bigl|\widetilde{x}^{(j)}([u])\bigr|^2
= \sum_{[u]\in \mathcal{V}^{(j-1)}} \frac{\bigl|\sum_{v\in [u]} x^{(j)}(v)\bigr|^2}{|[u]|}.
\end{align*}
Compared to the full wavelet transform $\lVert (\Phi^{(j)})^{\top} x^{(j)}\rVert^2 = \sum_{[u]\in \mathcal{V}^{(j-1)}} \sum_{v\in [u]} \bigl|x^{(j)}(v)\bigr|^2$, the compressive wavelet transform, while denoising the detail information, takes the average information of $x^{(j)}$ that belongs to the same cluster of level $j-1$.

The benefits of adopting the wavelet-based MRA pooling layers on GNN come from three perspectives. First, the compressive transformation purifies graph information adequately. Only the low-frequency coefficients are preserved to approximate the original data within each pooling layer, and the remaining noise in the high-frequency coefficients is stripped out. Second, as a result of performing local graph Fourier transformation, both the subgraph structure and the node features are involved in generating the supernodes' representation. As pointed out by \cite{zhang2018end, Morris2017Glocalized}, the compound information maximizes the model performance. Moreover, the characteristic function for Haar-like wavelet construction incorporates the chain's spatial behaviour in the spectral computation. As discussed above, the wavelet compression acts similar to the average pooling operation. Such a strategy is not only backed by theoretical novelty but also powerful in practical applications. 
\section{Experimental Study}
\label{experiment}
This section evaluates our proposed framework $\textsc{MathNet}$ on several graph classification and regression tasks. In Section~\ref{task1}, we select four popular graph classification data sets and compare the performance of our $\textsc{MathNet}$ with some representative methods in the literature. Also, we demonstrate the capability of learning on a large-scale data set for our proposed method with a multi-class classification task in Section~\ref{task2}. Finally, we present one regression task in Section~\ref{task3}. We present the partial descriptive statistics of all the data sets used in this paper in TABLE~\ref{partial_statistics} and provide detailed descriptions in the corresponding subsections.

All the programs in this work were written in PyTorch, and the library PyTorch Geometric \cite{pyg}. All the experiments run on NVIDIA\textsuperscript{\textregistered} Tesla V100 GPU with 5,120 CUDA cores, and 16GB HBM2 mounted on a high performance computing cluster.

\begin{table*}[th]
\caption{Statistical information of the data sets used for graph classification and regression tasks.}
\label{partial_statistics}
\begin{center}
\footnotesize{
\begin{tabular}{l c c c c c c c c}
\toprule
\multirow{1}{*}{\textbf{Data sets}} 
& \textbf{PROTEINS} & \textbf{ENZYMES} & \textbf{D\&D} & \textbf{MUTAG} & \textbf{PP(0.30)}$^{*}$ & \textbf{PP(0.35)}$^{*}$ & \textbf{PP(0.40)}$^{*}$ & \textbf{QM7}$^{**}$\\
\midrule
Max. \#Nodes & 620 & 126 & 5,748 & 28 & 1,000 & 1,000 & 1,000 & 23\\
Min. \#Nodes & 4 & 2 & 30 & 10 & 100 & 100 & 100 & 4\\
Avg. \#Nodes & 39.06 & 32.63 & 284.32 & 17.93 & 478 & 474 & 475 & 15.44\\
Avg. \#Edges & 72.82 & 62.14 & 715.66 & 19.79 & 1632.50 & 1611.50 & 1610.00 & 122.83\\
\#Graphs & 1,113 & 600 & 1,178 & 188 & 15,000 & 15,000 & 15,000 & 7,165\\
\#Classes & 2 & 6 & 2 & 2 & 3 & 3 & 3 & 1\\
\bottomrule
\multicolumn{9}{l}{$^{*}$ The number included in the bracket indicates the type of the \textbf{P}oint\textbf{P}attern data set, and the level of difficulty of the}\\
\multicolumn{9}{l}{\hspace{0.3cm}task. The meaning of this number will be explained in Section~\ref{task3}.}\\
\multicolumn{9}{l}{$^{**}$ The data set is used for the regression task and the \#Classes represents number of regression targets.}
\end{tabular}}
\end{center}
\end{table*}

\subsection{Experimental Setup}
\label{exp_setup}
We tune the model architecture of our proposed $\textsc{MathNet}$ to produce the best performance for each data set. The architecture search is conducted based on the number of pooling operations and the number of convolutional layers before each pooling operation. A two- or three-layer multilayer perceptron (MLP) is used as the classifier after the sequence of convolutional and pooling layers across all the experiments. The best architecture of $\textsc{MathNet}$ for each data set is reported in the corresponding subsections. We employ spectral clustering \cite{shi2000normalized, stella2003multiclass} to generate the coarse-grained chain with a given number of tree layers which depends on the number of pooling layers required in the network. The number of parents/clusters in each coarsened level is set to half of the number of nodes in the finer layer. We force the top layer of the tree to be clustered into a single cluster which corresponds to the unified vectorial graph representation. Spectral clustering has a provable capability to cluster various data patterns and handle the graph with isolated nodes, which only requires the graph structure to be the input argument of the algorithm.

We split the data set in each experiment into training, validation and test sets by 80\%, 10\% and 10\%, respectively. As noticed in \cite{shchur2018pitfalls}, different data splits will affect a GNN model's performance to a great extent. Thus, we repeat each experiment 10 times with random shuffling for the data set before splitting. Note that there is also randomness involved in the spectral clustering step, which leads to a slightly different coarse-grained chain generated for each graph. To rule out this unwanted randomness, we pre-process each data set and attach the generated coarse-grained chain with associated Haar bases. Therefore, the random shuffling will not affect the structures of the coarse-grained chains.

We report the mean accuracy (or loss in the regression task) and the standard deviation of our model in all experiments. For the baseline methods in Section~\ref{task1}, unless otherwise specified, we only report the best-published results from the original papers of the available data sets. For Section~\ref{task2} and~\ref{task3}, we will specify the sources where the results of the baseline methods are retrieved.

We use the Adam optimizer \cite{adam_optimizer} with an early stopping criterion as suggested in \cite{shchur2018pitfalls}. Specifically, we set a maximum of 150 epochs and stop training if the validation loss does not improve for consecutive 25 epochs. For the tasks in Sections~\ref{task2} and~\ref{task3}, we turn off the early stopping, and the experiment runs for max epoch 20 and 100. We use a simple grid search based on the training and validation sets to perform the hyperparameter tuning. We document a list of the hyperparameters of the model in TABLE~\ref{grid_search} along with their search spaces.

\begin{table}[th]
\caption{Grid search space for the hyperparameters used in all the experiments.}
\begin{center}
\begin{tabular}{l c}
\toprule
\textbf{Hyperparameter} & \textbf{Choice}\\
\midrule
Learning rate & 1$\mathrm{e}$-2, 5$\mathrm{e}$-3, 1$\mathrm{e}$-3, 5$\mathrm{e}$-4, 1$\mathrm{e}$-4\\
Hidden size & 16, 32, 64, 128\\
Weight decay (L2) & 5$\mathrm{e}$-3, 1$\mathrm{e}$-3, 5$\mathrm{e}$-4, 1$\mathrm{e}$-4\\
Batch size & 32, 64, 128, 256\\
\#Poolings & 1, 2, 3\\
\bottomrule
\end{tabular}
\label{grid_search}
\end{center}
\end{table}

\subsection{Graph Classification Benchmarks}
\label{task1}
In this part, we test the proposed $\textsc{MathNet}$ on four graph classification benchmarks, and compare its performance with traditional graph kernel methods and recently developed GNN models. 

We give the selected benchmark data sets as follows. \textbf{D\&D} \cite{dobson2003distinguishing, shervashidze2011weisfeiler} is a protein graph data set consisting of a collection of protein structures. A graph represents each protein in the data set. The nodes are amino acids, and an edge connects two nodes if they are less than six angstroms apart. The node features of each graph are the binary encoding of some chemical properties. The task is to classify the protein structures into enzymes and non-enzymes; \textbf{PROTEINS} \cite{dobson2003distinguishing, borgwardt2005protein} is also a protein structure data set which we consider as a simplified version of \textbf{D\&D} with the same task. The protein structures documented in \textbf{PROTEINS} are much smaller than in \textbf{D\&D} in terms of the node and edge numbers. \textbf{MUTAG} \cite{debnath1991structure, kriege2012subgraph} is a mutagen data set comprising 188 compounds. Each compound is represented by a graph where atoms are nodes, and covalent bonds are edges. The task is to predict whether a compound in the data set is mutagenic based on its compound structure and chemical properties. \textbf{ENZYMES} \cite{borgwardt2005protein, schomburg2004brenda} is a graph data set consisting of 100 proteins. The proteins will be classified into six Enzymes Commission top-level enzyme classes (EC classes), such as oxidoreductases (EC1) and transferases (EC2). It is thus a multi-class graph classification task.

To demonstrate the performance of our proposed $\textsc{MathNet}$ on these four benchmark classification data sets, we consider the following state-of-the-art methods as the baselines:
\begin{itemize}
    \item \textsc{GIN} \cite{xu2018how} generalizes the Weisfeiler-Lehman graph isomorphism test and thus achieves maximum discriminative power among the class of GNNs.
    \item \textsc{PatchySan} \cite{niepert2016learning} uses a receptive field on nodes to extracts locally connected regions of graphs.
    \item \textsc{DGCNN} \cite{zhang2018end} proposes the SortPooling layer with convolution on the sorted graph nodes.
    \item \textsc{DiffPool} \cite{ying2018hierarchical} learns a soft assignment matrix by a separate pooling GNN module. The assignment matrix is then used to transform the node embeddings and the graph structure.
    \item \textsc{SAGPool} \cite{lee2019self} develops a pooling layer based on self-attention mechanism using graph convolution. \textsc{SAGPool} takes into account both the node feature and graph structure.
    \item \textsc{EigenPool} \cite{ma2019graph} is a spectral pooling method which relies on graph Fourier transforms. They employ the traditional GCN \cite{kipf2016semi} for graph convolution and form a complete framework for graph representation learning.
    \item \textsc{g-U-Nets} \cite{gao2019graph} is an encoder-decoder architecture which involves the proposed graph pooling (gPool) and unpooling (gUnpool) procedures.
    \item We also compare with several graph kernel methods, including Shortest-Path kernel (\textsc{SP}) \cite{borgwardt2005shortest}, \textsc{Graphlet} count kernel \cite{shervashidze2009efficient}, random walk kernel (\textsc{RW}) \cite{gartner2003graph} and Weisfeiler-Lehman subtree kernel (\textsc{WL}) \cite{shervashidze2011weisfeiler}.
\end{itemize}

The model architectures of our $\textsc{MathNet}$ used in \textbf{PROTEINS} and \textbf{MUTAG} are two Haar convolutional (HaarConv) layers followed by one Haar pooling (HaarPool) layer and a three-layer MLP. The architecture used in \textbf{D\&D} and \textbf{ENZYMES} are two HaarConv layers followed by one HaarPool layer and a two-layer MLP as the classifier. For each of these tasks, we apply batch normalization \cite{ioffe2015batch} for all layers of MLP, except for the output layer.

We summarize the results in TABLE~\ref{task1_results}. Overall, our proposed $\textsc{MathNet}$ obtains a superior performance against the baselines on all the data sets. For \textbf{PROTEINS} and \textbf{D\&D}, $\textsc{MathNet}$ achieves the top test accuracy. For \textbf{ENZYMES} and \textbf{MUTAG}, the performance of our $\textsc{MathNet}$ is ranked top-three among all the baselines. It demonstrates that our proposed model can effectively extract global topological information of input graph via a hierarchical learning process, and can adequately capture latent node clustering information by the chain-based sparse Haar representation.

\begin{figure}[th]
\centering
\begin{subfigure}{0.48\textwidth}
    \includegraphics[width=0.95\linewidth]{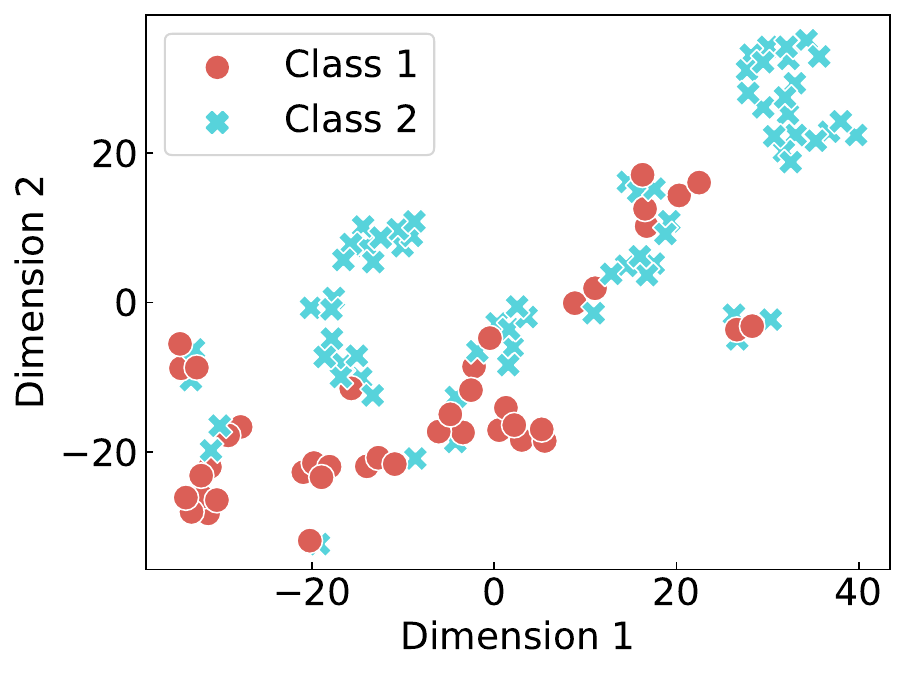}
\end{subfigure}
\hfill
\begin{subfigure}{0.48\textwidth}
    \includegraphics[width=0.95\linewidth]{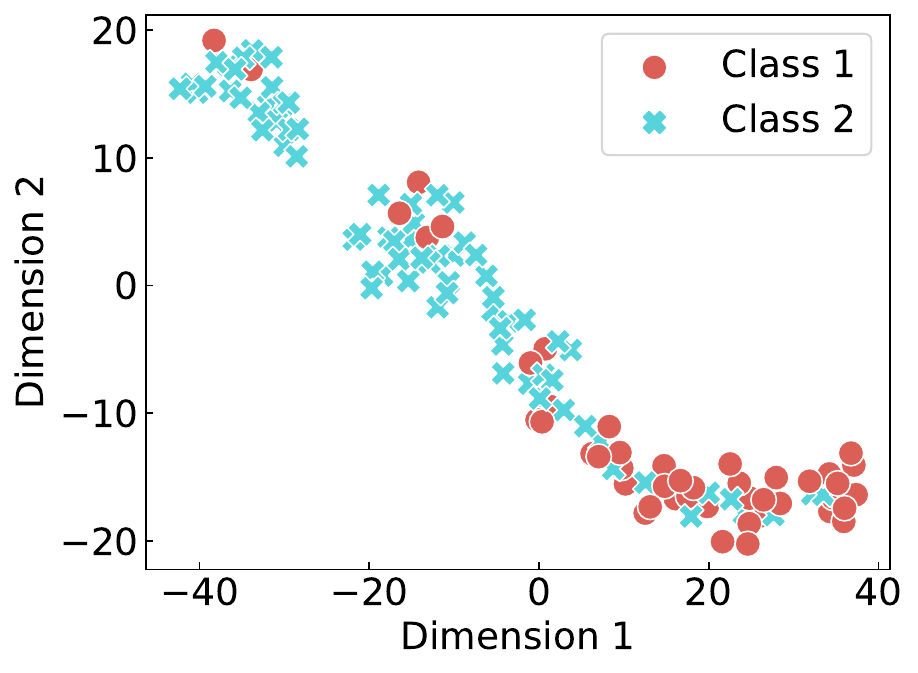}
\end{subfigure}
\caption{t-SNE visualizations on test set. The points denote the graphs, and the different colors represent the true class labels. Left: \textbf{PROTEINS}. Right: \textbf{D\&D}.}
\label{fig:tsne}
\end{figure}

Moreover, we exploit the t-distributed Stochastic Neighbor Embedding (t-SNE) to visualize the test classification results of the pre-trained classifier on \textbf{PROTEINS} and \textbf{D\&D}. (We exclude \textbf{ENZYMES} and \textbf{MUTAG} due to the limited number of test samples in these two data sets.) We use a two-dimensional embedding space for visualization, and the point in the embedding space corresponds to a graph of the test set. Fig.~\ref{fig:tsne} suggests that there is a visible clustering pattern of the graphs processed by the sequence of HaarConv and HaarPool layers in our model.

\begin{table}[th]
\caption{Mean test accuracy (in percentage) and standard deviation of $\textsc{MathNet}$ as compared with existing models on benchmark graph classification data sets, over 10 repetitions.}\label{task1_results}
\begin{center}
\setlength{\tabcolsep}{15pt}
\begin{tabular}{l c c c c}
\toprule
\textbf{Methods} & \textbf{PROTEINS} & \textbf{ENZYMES} & \textbf{D\&D} & \textbf{MUTAG} \\
\midrule
\textsc{SP} & 75.07$^*$ & 42.32\textsuperscript{a} & -- & 85.79$^*$ \\
\textsc{Graphlet} & 71.67$^*$ & 41.03\textsuperscript{a}  & 78.45$^*$ & 81.58$^*$ \\
\textsc{RW} & 74.22$^*$ & -- & -- & 83.68$^*$ \\
\textsc{WL} & 72.92$^*$ & 53.43\textsuperscript{a} & 77.95$^*$ & 80.72$^*$\\
\midrule
\textsc{GIN} & 76.2 & -- & -- & \textcolor{blue}{89.4} \\
\textsc{PatchySan} & 75.00 & -- & 76.27 & \textbf{91.58} \\
\textsc{DGCNN} & 75.54 & 57.12\textsuperscript{a} & 79.37 & 85.83 \\
\textsc{DiffPool} & 76.25 & \textcolor{red}{62.53}\textsuperscript{a} & \textcolor{blue}{80.64} & -- \\
\textsc{SAGPool} & 72.17 & -- & 77.07 & -- \\
\textsc{EigenPool} & \textcolor{blue}{76.6} & \textbf{65.0} & 78.6 & -- \\
\textsc{g-U-Nets} & \textcolor{red}{77.68} & -- & \textcolor{red}{82.43} & -- \\
\midrule
\textbf{$\textsc{MathNet}$} & \textbf{78.3$\pm$1.60} & \textcolor{blue}{62.5$\pm$3.85} & \textbf{82.5$\pm$3.59} & \textcolor{red}{89.6$\pm$2.49}\\ 
\bottomrule
\multicolumn{5}{l}{`$*$' denotes the record retrieved from \textsc{PatchySan} \cite{niepert2016learning}.}\\
\multicolumn{5}{l}{`a' denotes the record retrieved from \textsc{DiffPool} \cite{ying2018hierarchical}.}\\
\multicolumn{5}{l}{ `--' means that there is no public
record for the method on the dataset.}\\
\multicolumn{5}{l}{! The records without superscription are from their corresponding original papers.}\\
\multicolumn{5}{l}{! The decimal place is not modified when transferring the results.}\\
\multicolumn{5}{l}{! The \textbf{best} result is in \textbf{bold}, and the \textcolor{red}{second} and \textcolor{blue}{third} positions are marked in \textcolor{red}{red} and}\\
\multicolumn{5}{l}{\hspace{0.3cm}\textcolor{blue}{blue}, respectively.}
\end{tabular}
\end{center}
\end{table}

\subsection{Large Data Set for Point Distribution Recognition}
\label{task2}
There are many benchmark graph data sets used in the literature for graph classification, see \cite{dwivedi2020benchmarking, hu2020open} for examples. However, many data sets still suffer from not understanding their underlying mechanism, which may lead to a learning method generating an unexplainable graph representation.

We introduce a novel large-scale graph classification data set composed of simple point patterns from the statistical mechanics. We simulate three types of point patterns in two-dimensional space: hard disks in equilibrium (HD), Poisson point process (PPP) and random sequential adsorption (RSA) of disks. The PPP and HD are the models that are typically used to describe the micro-structures of gases and liquids \cite{hansen1990theory}. The RSA is a non-equilibrium stochastic process which simulates the particles one by one with some satisfied non-overlapping conditions. The point patterns simulated by these three mechanics are structurally different; thus, a collection of these point patterns forms a graph classification task with the three simulation models being the class labels. Each point pattern is represented as a graph, where the particles are reviewed as nodes, and there is an edge connecting two nodes if the two particles are within a threshold distance. The node degree is used as the feature for each node of all the graphs. We name the resulting data set as \textbf{PointPattern}. Compared to the data sets used in Section~\ref{task1}, \textbf{PointPattern} is large-scale in terms of sample size, graph size, and the number of connections within each graph.

The volume fraction covered by particles $\phi_{\textnormal{HD}}$ of HD is fixed at 0.5, and this factor for the point pattern simulated from PPP is $\phi_{\textnormal{PPP}}=0$. Moreover, we can tune the corresponding factor $\phi_{\textnormal{RSA}}$ of the RSA model to control the similarity between RSA and the other two simulation models. If $\phi_{\textnormal{RSA}}$ becomes closer to 0.5, it is harder to distinguish the point patterns RSA from HD. Hence, the factor $\phi_{\textnormal{RSA}}$ can be used to adjust the level of difficulty of the task. In this experiment, we consider three data sets with varying levels of difficulty $\phi_{\textnormal{RSA}}\in\{0.30, 0.35, 0.40\}$.

\begin{figure}[ht]
\centering
\begin{subfigure}{0.32\textwidth}
    \includegraphics[width=0.95\linewidth]{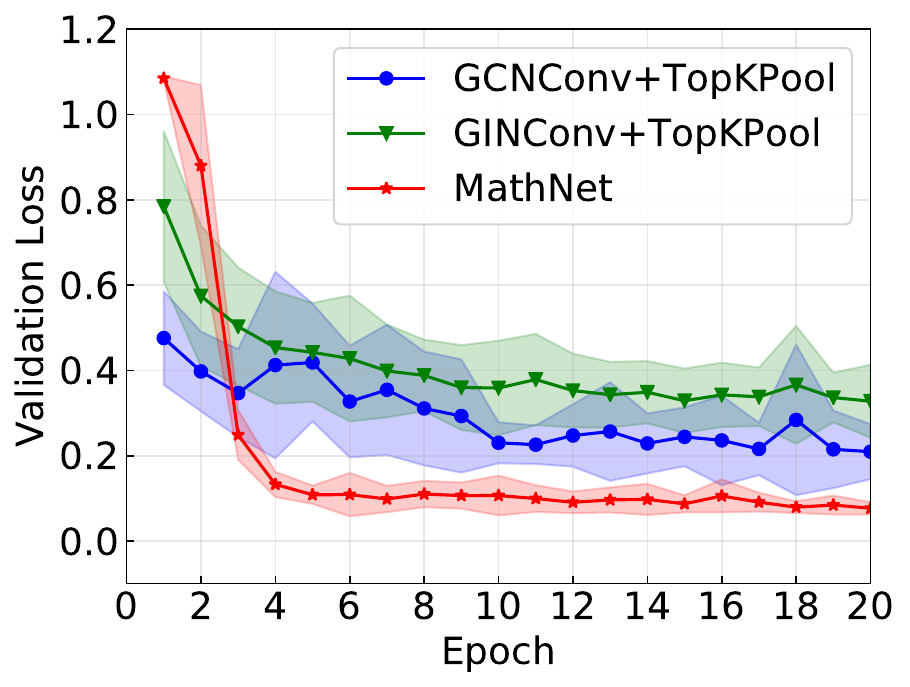}
\end{subfigure}
\begin{subfigure}{0.32\textwidth}
    \includegraphics[width=0.95\linewidth]{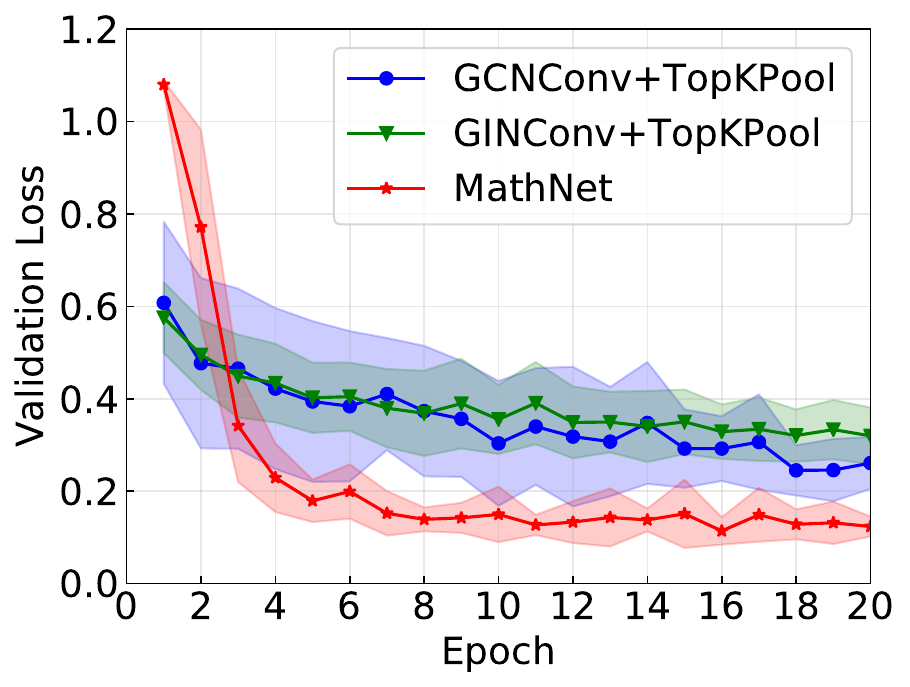}
\end{subfigure}
\begin{subfigure}{0.32\textwidth}
    \includegraphics[width=0.95\linewidth]{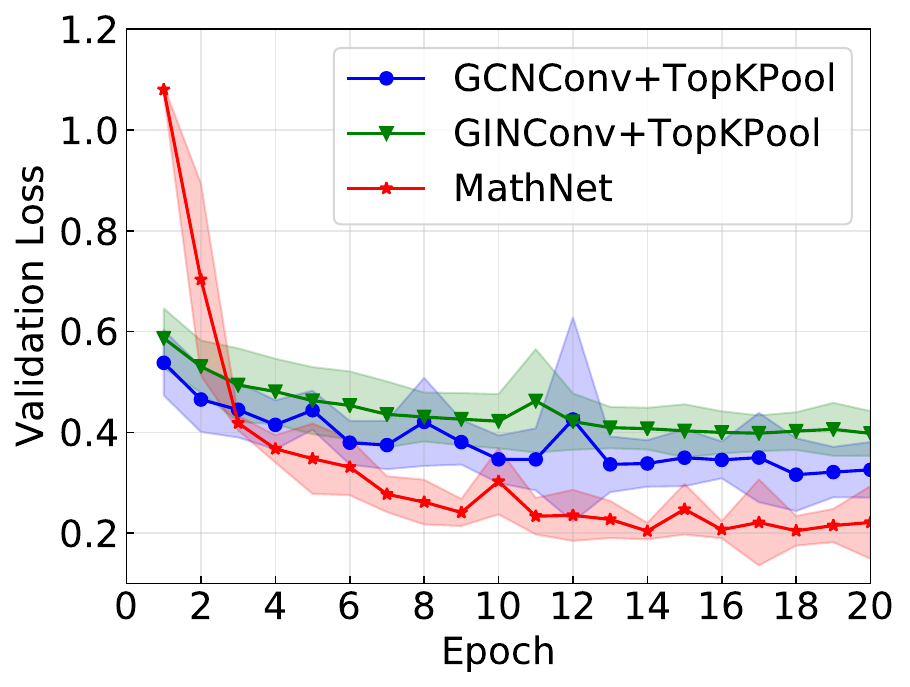}
\end{subfigure}

\medskip
\begin{subfigure}{0.32\textwidth}
    \includegraphics[width=0.95\linewidth]{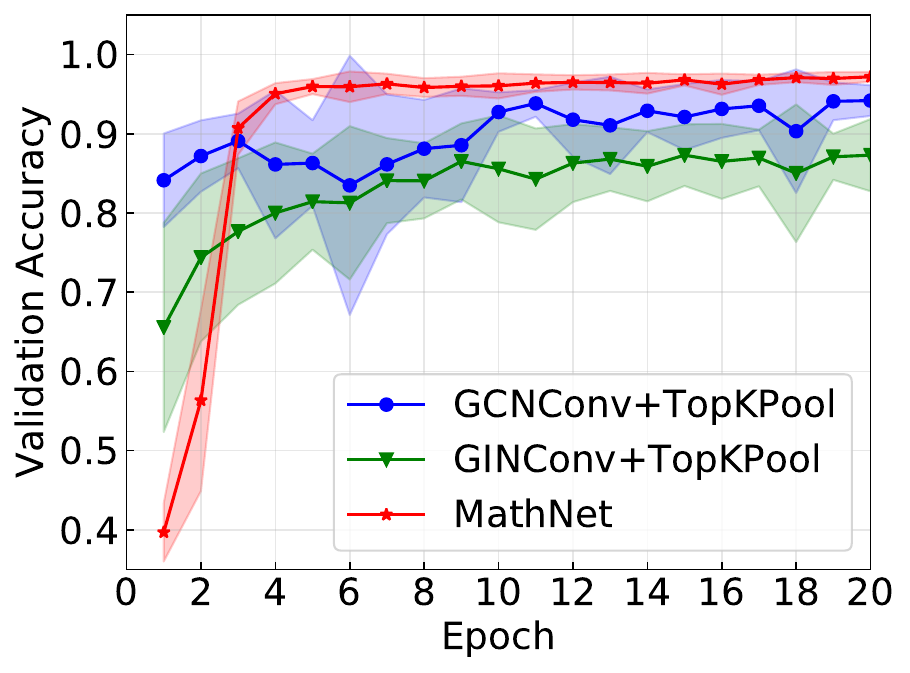}
    \caption{$\phi_{\textnormal{RSA}} = 0.30$}
\end{subfigure}
\begin{subfigure}{0.32\textwidth}
    \includegraphics[width=0.95\linewidth]{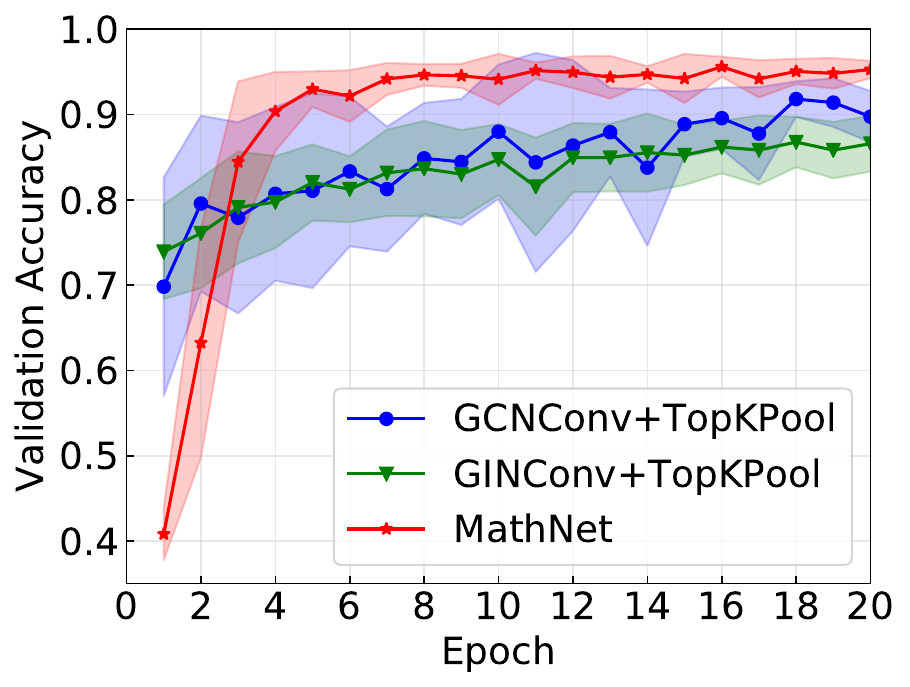}
    \caption{$\phi_{\textnormal{RSA}} = 0.35$}
\end{subfigure}
\begin{subfigure}{0.32\textwidth}
    \includegraphics[width=0.95\linewidth]{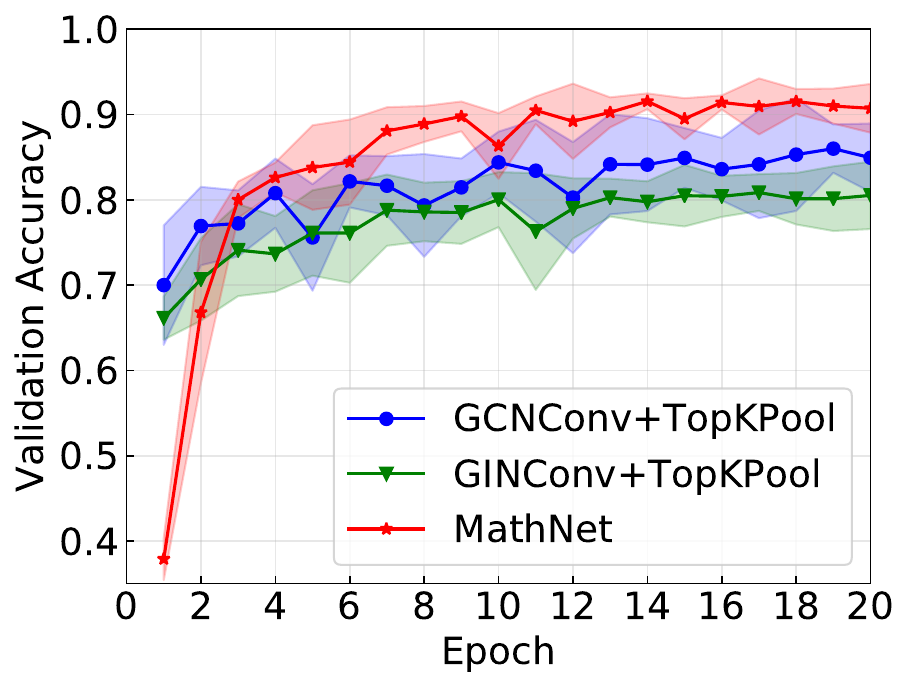}
    \caption{$\phi_{\textnormal{RSA}} = 0.40$}
\end{subfigure}
\vskip 2mm
\caption{The plots of validation loss against epoch and validation accuracy against epoch for \textbf{PointPattern} with three different levels of difficulty $\phi_{\textnormal{RSA}}\in\{0.30, 0.35, 0.40\}$.}
\label{fig:pp_plot}
\end{figure}

We compare our proposed $\textsc{MathNet}$ with two GNN baselines: \textsc{GCNConv+TopKPool} uses GCN \cite{kipf2016semi} convolutional layer with TopK \cite{gao2019graph,cangea2018towards} pooling layer; \textsc{GINConv+TopKPool} employs GIN \cite{xu2018how} convolutional layer with TopK %\cite{gao2019graph,cangea2018towards} 
pooling layer. All the models, including our $\textsc{MathNet}$ share the same neural architecture: three units of alternating convolutional layer and pooling layer followed by a three-layer MLP. We apply the dropout \cite{hinton2012improving} only for the first layer of MLP in each model to prevent from over-fitting. In \textsc{GCNConv+TopKPool} and \textsc{GINConv+TopKPool}, we use global max pooling to unify the graph representation before the MLP classifier. In this experiment, we fix the number of hidden neurons to 64, learning rate to 0.001 and weight decay to 0.0005 for all the models. Each \textbf{PointPattern} data set is a 3-classification task on 15,000 graphs (5,000 for each class) with graph node size varying between 100 and 1,000. We refer the reader to TABLE~\ref{partial_statistics} for more statistics of the data sets. The data split and other experimental settings are the same as the procedures described in Section~\ref{exp_setup}. The experimental results are reported in TABLE~\ref{task2_results} and Fig.~\ref{fig:pp_plot}.

\begin{table*}[t]
\caption{Mean test accuracy (in percentage) and standard deviation of $\textsc{MathNet}$ and the other two baseline models on \textbf{PointPattern} with three different levels of difficulty $\phi_{\textnormal{RSA}}$. The test accuracy is averaged over 10 repetitions each with 20 epochs and different random seeds.}
\begin{center}
\begin{tabular}{c c c c}
\toprule
\textbf{PointPattern} & \textsc{GINConv + SAGPool} & \textsc{GCNConv + TopKPool} & $\textsc{MathNet}$\\
\midrule
$\phi_{\textnormal{RSA}}=0.30$ & 90.9$\pm$2.95 & 92.9$\pm$3.21 & \textbf{97.4$\pm$0.34}\\
$\phi_{\textnormal{RSA}}=0.35$ & 86.7$\pm$3.30 & 89.3$\pm$3.31 & \textbf{96.0$\pm$0.59}\\
$\phi_{\textnormal{RSA}}=0.40$ & 80.2$\pm$3.80 & 85.1$\pm$4.06 & \textbf{92.7$\pm$0.72}\\
\bottomrule
\end{tabular}
\label{task2_results}
\end{center}
\end{table*}

From the results in TABLE~\ref{task2_results}, we can see that our $\textsc{MathNet}$ outperforms the baselines \textsc{GCNConv+TopKPool} and \textsc{GINConv+TopKPool} on each of three data sets by a large margin with 5 to 7 percentage points higher mean test accuracy. Moreover, $\textsc{MathNet}$ also achieves much smaller standard deviations than the baseline methods across all the data sets. We further compare the trends of validation loss and accuracy against epoch of the proposed $\textsc{MathNet}$ to those of the baselines in Fig.~\ref{fig:pp_plot}. Although $\textsc{MathNet}$ has a higher validation loss during the first 2 epochs than the baseline methods, it converges rapidly and reaches the stationary point within 7 epochs in training each data set. Fig.~\ref{fig:pp_plot} also shows that $\textsc{MathNet}$ obtains much smaller variations in terms of validation loss and validation accuracy. The lower validation loss, higher validation accuracy and higher test accuracy of our $\textsc{MathNet}$ on all the data sets indicate its superior learning and generalization abilities. Based on these observations, we can conclude that the proposed $\textsc{MathNet}$ is a highly effective and robust model for large-scale graph classification tasks.

\subsection{Graph Regression}
\label{task3}
In this part, we evaluate the performance of the $\textsc{MathNet}$ on a graph regression task with \textbf{QM7} data set \cite{blum, rupp} from the field of quantum chemistry. The data set consists of 7,165 molecules, each of which contains up to 23 atoms. We represent each molecule by a graph with the atoms as nodes and bonds as edges. Then, the Coulomb energy matrix of each molecule becomes the adjacency matrix which depicts the topological structure of the corresponding graph. The atomization energy value of the molecule is the regression label. Since the atom itself is featureless, we create an uninformative feature (i.e., constant scalar 1) for each node of all the graphs. Under this construction, the regressor model will only leverage the structure of the graphs.

In the experiment, we standardize the target values for the training procedure and then convert the predicted value back to the original domain for evaluating the model on the validation and test sets. As suggested in \cite{gilmer2017neural}, we use mean squared error (MSE) as the metric for the training and mean absolute error (MAE) for the evaluations. To validate the capability of our model, we borrow the following baselines along with their records directly from \cite{wu2018moleculenet} for comparison: Random Forest (\textsc{RF}) \cite{breiman2001random}, Multitask Networks (\textsc{Multitask}) \cite{ramsundar2015massively}, Kernel Ridge Regression (\textsc{KRR}) \cite{cortes1995support}, and Graph Convolutional models (\textsc{GC}) \cite{altae2017low}. To make more comparison, we test on the model with convolution plus pooling unit \textsc{GCNConv+SAGPool}. We employ one \textsc{GCN} \cite{kipf2016semi} convolutional layer followed by one widely used graph pooling layer \textsc{SAGPool} \cite{lee2019self} and a three-layer multilayer perceptron (MLP) as the classifier. The neural architecture of our $\textsc{MathNet}$ is identical to the baseline \textsc{GCNConv+SAGPool}.

\begin{table}[ht]
\caption{Average of test mean absolute error (MAE) and standard deviation over 10 repetitions on \textbf{QM7} for the graph regression task.}
\begin{center}
\begin{tabular}{c | c}
\toprule
\textbf{Methods} & \textbf{Test MAE}\\
\midrule
\textsc{RF} & 122.7$\pm$4.2$^*$\\
\textsc{Multitask} & 123.7$\pm$15.6$^*$\\
\textsc{KRR} & 110.3$\pm$4.7$^*$\\
\textsc{GC} & 77.9$\pm$2.1$^*$\\
\midrule
\textsc{GCNConv+SAGPool} & 43.6$\pm$0.98\\
% GCN-HaarPool & 42.9$\pm$1.2\\
\midrule
\textbf{$\textsc{MathNet}$} \textbf{(Ours)} & \textbf{42.7$\pm$0.92}\\
\bottomrule
\multicolumn{2}{l}{`$*$' indicates the records retrieved from \cite{wu2018moleculenet}.}
\end{tabular}
\label{task3_results}
\end{center}
\end{table}

TABLE~\ref{task3_results} shows the results of the experiment. The $\textsc{MathNet}$ achieves the lowest mean test MAE among all the baselines, which demonstrate the superior performance of our proposed method on this regression task. Compared to the baseline \textsc{GCNConv+SAGPool} model, $\textsc{MathNet}$ not only obtains an even lower mean test MAE but also has a slightly smaller standard deviation (calculated in the same number of repetitions and epochs). These observations confirm that our proposed $\textsc{MathNet}$ is an effective and robust model for graph regression task.

\FloatBarrier
\section{Conclusion}
\label{conclusion}
We proposed \textsc{MathNet} for graph classification and regression tasks. It is an end-to-end graph neural network with interrelated graph convolutional and graph pooling layers. The model learns efficient graph representation in the convolution and unified topological information in the pooling. Our method establishes a general framework for fast computation for GNN with any Haar-like MRA wavelets in comparison to existing approaches. By decomposing the input graph signals to multiresolution, both local and global information from the vertex domain are preserved, and the extracted mutual information among graphs are fully researched. Novel theories of signal processing back up this spectral graph framework. This MRA-based graph model with a Haar-like wavelet, though with the simplest choice of a uniform `father wavelet', achieves superior prediction accuracy and stable performance in extensive experiments. Numerical evidence proves that our method outperforms other GNN methods, especially on large-scale data sets.

We left some potentials for future study. First, the expressivity power of the model is worth exploring. While empirical results have shown its capability, it is worth investigating how selecting different wavelets would impact the model performance. A rigorous theoretical analysis on the expressiveness of \textsc{MathNet} could help solidify the usefulness of our framework and potentially for more general spectral-based methods. Although we test our model mostly on the well-explored benchmark data sets, it could also be useful for real-world applications, such as building knowledge graphs for drug repositioning for COVID-19. Besides, the design of graph clustering needs to investigate further. It is not yet clear to what extent the clustering methods would impact the performance of our proposed \textsc{MathNet}, and what is the guideline for choosing the clustering methods that could better capture graph topological information. Since the coarse-grained chain is built upon the clustering results, methods that encode specific edge-connection patterns could be an extra benefit to our model.

%\section*{References}
\bibliography{reference}

\end{document}